\documentclass[letterpaper, 10 pt, conference]{config/ieeeconf}  %

\IEEEoverridecommandlockouts                              %

\usepackage{etoolbox}
\makeatletter 
\let\NAT@parse\undefined
\makeatother
\usepackage[numbers]{natbib} 

\usepackage{multirow}
\usepackage[bookmarks=true, colorlinks=false,
    linkcolor=black,
    filecolor=magenta,      
    urlcolor=cyan]{hyperref}
\usepackage{url}
\usepackage{graphicx} %
\usepackage{amsfonts,amssymb,amsmath} 
\usepackage{graphics} 
\usepackage{epsfig} 
\usepackage[all]{xy}  
\usepackage{algorithm}
\usepackage{algpseudocode}

\usepackage[labelfont=small,textfont={small}]{caption}
\usepackage{subcaption}
\usepackage{epstopdf} 
\usepackage{balance}
\usepackage{soul}
\usepackage{color}
\usepackage{booktabs}   %
\usepackage{siunitx}    %
\usepackage{caption}
\newtheorem{problem}{Problem}

\DeclareMathOperator{\diag}{diag}
\usepackage[dvipsnames]{xcolor}
\usepackage{pgf}
\usepackage{adjustbox}
\usepackage{pgffor}
\usepackage{etoolbox}
\usepackage{tabularx}
\usepackage{gensymb}
\usepackage{pifont}
\usepackage{pgfplotstable}
\usepackage{tikz}
\usepackage{pgfplots}
\pgfplotsset{
compat=1.17,
}
\usepgfplotslibrary{statistics}
\usetikzlibrary{pgfplots.groupplots,pgfplots.statistics } %
\usetikzlibrary{arrows.meta, positioning, shapes.geometric}

\title{\LARGE 

Localization in Spatiotemporal Fields via Environmental PDEs

}

\author{Jose Fuentes$^{1}$, Abdullah Al Redwan Newaz$^{2}$, Ana Cavalcanti$^{1}$ and Leonardo Bobadilla$^{1}$%
\thanks{$^{1}$J. Fuentes, A. Cavalcanti, and L. Bobadilla are with the Faculty of Computer  Engineering and Information Sciences, Florida International University, Miami, FL 33174, USA.
{\tt\small jfuen099@fiu.edu, acavalca@fiu.edu, jabobadi@fiu.edu}.
$^{2}$A. Redwan is with the University of New Orleans, New Orleans, LA 70148, USA,
{\tt\small aredwann@uno.edu}. 
This work is supported in part by the U.S. EPA grant BR-02F47801-
5010M, NSF grants 2118329, IIS-2024733, IIS-2331908, ONR grant
N00014-23-1-2789, the DoD grant 78170-RT-REP, the ARL under contract
W911NF1920243. This is contribution \#2161 from the Institute of Environment at Florida International University.
}
}

\begin{document}
\setlength{\textfloatsep}{3pt}  %
\setlength{\floatsep}{3pt}      %
\setlength{\intextsep}{3pt}
\maketitle
\thispagestyle{empty}
\pagestyle{empty}
\maketitle              %

\begin{abstract}
     This paper proposes a localization framework that uses spatiotemporal fields governed by partial differential equations (PDEs) as localization signatures. Two PDE classes are considered: the shallow water equations, which describe free-surface flows in coastal and riverine environments, and the advection-diffusion equation, which models the transport and mixing of scalar quantities such as temperature, salinity, and dissolved oxygen. A numerical PDE solver provides predicted fields over the domain, and multiple field channels are fused as multimodal measurements to improve localization accuracy. We formulate the problem within a Rao-Blackwellized particle filter (RBPF) that partitions the vehicle state into a nonlinear component sampled by particles and a linear sensor bias component tracked analytically via per-particle Kalman filters. This factorization reduces the required number of particles compared to a standard particle filter while accounting for realistic sensor drift. Simulation studies on both PDE scenarios show that the RBPF consistently outperforms a standard particle filter in terms of final position error and Root Mean Square Error (RMSE) across varying particle counts. Field experiments with an autonomous surface vehicle measuring salinity, temperature, and dissolved oxygen validate that PDE-governed environmental fields provide sufficient spatial variability for practical localization. Related experimental videos are available at \url{https://localization-environmental-pdes.github.io/}.
\end{abstract}

\section{Introduction}\label{sec:introduction}

Localization in GPS-denied environments is a fundamental challenge for autonomous vehicles operating underwater, indoors, or in contested electromagnetic settings. In such scenarios, the vehicle cannot rely on satellite navigation and must instead exploit alternative information sources. Prior work has considered magnetic field anomalies~\cite{solin2018modeling, solin2016terrain}, radio signal strength~\cite{ferris2007wifi}, and visual landmarks~\cite{montemerlo2002fastslam} as substitutes. A common thread in these approaches is that a spatially varying physical field is matched against onboard sensor readings to constrain the vehicle position. However, many physical fields of practical interest, such as those arising in ocean surveying, pollution tracking, and hazard detection~\cite{dunbabin2012robots}, exhibit significant temporal dynamics in addition to spatial structure. Relying on a purely spatial representation ignores temporal correlations that, if modeled, could sharpen localization estimates and extend the range of usable environmental signatures. A spatiotemporal modeling framework is therefore required to capture both the spatial variability and the temporal evolution of such fields.

A small but growing body of work has begun to address localization and mapping in fluid flow environments. In~\cite{song2014towards,song2019flam} and~\cite{kumar2025multiflow}, flow-based localization and mapping frameworks were introduced that exploit flow field structure for simultaneous state and map estimation. However, these methods are restricted to steady or slowly varying flows, or to specific dynamical models amenable to data-driven decomposition. They cannot accommodate the full range of spatiotemporal fields encountered in environmental applications, particularly those governed by the advection-diffusion equation or the shallow water equations, which exhibit complex wave propagation, diffusive spreading, and nonlinear coupling.

\begin{figure}
    \centering
\resizebox{\linewidth}{!}{
\begin{tikzpicture}[node distance=2cm, auto]
    \node (initial) {\includegraphics[width=0.4\textwidth]{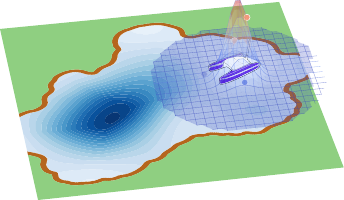}};
    \node (uncertain) [right=of initial] {\includegraphics[width=0.4\textwidth]{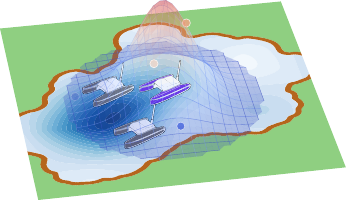}};
    \node (fields) [below=1cm of uncertain] {\includegraphics[width=0.4\textwidth]{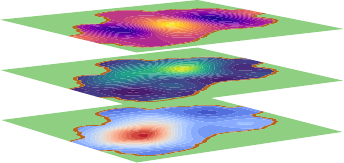}};
    \node (corrected) [left=of fields] {\includegraphics[width=0.4\textwidth]{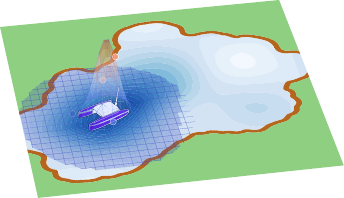}};

    \draw [-{Stealth[scale=1.5]}, line width=1.5pt] (initial) -- node[above] {Motion} (uncertain);
    
    \draw [-{Stealth[scale=1.5]}, line width=1.5pt] (fields) -- node[above, align=center] {Particle\\Correction} (corrected);
    
    \draw [-{Stealth[scale=1.5]}, line width=1.5pt] (uncertain) -- node[right] {Field measurement $\boldsymbol{\phi}$} (fields);

    \draw [-{Stealth[scale=1.5]}, line width=1.5pt] (corrected) -- node[align=center] {Next State} (initial);

\end{tikzpicture}
}
\caption{General localization scheme with a particle filter. The robot, when it moves, is less certain about its state (represented by the Gaussian surfaces). When it queries the environmental fields, it can correct its current location.}
    \label{fig:general_scheme}
    \end{figure}
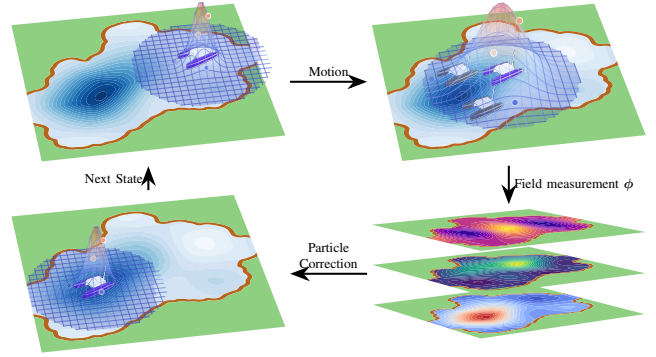

A separate line of research has investigated flow field estimation, adaptive navigation, and source identification~\cite{newaz2016uav} in environmental scalar fields. In~\cite{li2024enkode}, an active learning framework for unknown flows was proposed using Koopman operators and ensemble methods. In~\cite{chen2025mdcpp}, multi-robot dynamic coverage path planning with workload adaptation was addressed. In~\cite{jouffroy2004underwater, Fuentes2022}, partial differential equation (PDE) based observers are used to estimate underwater vehicle trajectories. In~\cite{fuentes2024adaptive}, a learning-based adaptive navigation framework for scalar field mapping was introduced, employing the advection-diffusion PDE and Gaussian Process Regression to estimate fields and their gradients in real time. In~\cite{wiedemann2017gas}, a probabilistic framework for multi-robot gas source localization was developed, modeling gas dispersion with the 2D diffusion PDE and using factor graph inference with Sparse Bayesian Learning to estimate source locations and magnitudes. While these works demonstrate the value of PDE-based environmental models for estimation and planning tasks, they all focus on flow modeling, coverage optimization, or source parameter estimation. None of them addresses the inverse problem of localizing the vehicles themselves, given field measurements in those realistic environments.

In the context of vehicle localization, the Rao-Blackwellized particle filter (RBPF) exploits conditional independence structure to factorize the posterior distribution into components estimated by particles and components updated analytically, yielding dramatic reductions in the required number of particles~\cite{doucet2000rbpf}. This technique has been successfully applied across diverse domains, yet it has not been extended to spatiotemporal field-based localization.
In~\cite{kok2024rbpf_slam} developed RBPF-based particle smoothing for SLAM with conditionally linear or linearized models. However, the maps considered are static, and the approach does not address spatiotemporal dynamics governed by environmental PDEs. In~\cite{shen2016connected} applied RBPF to connected vehicle localization but relies on GNSS signal measurements rather than environmental field observations and does not consider spatiotemporal field dynamics. In~\cite{taguchi2010rbpf} employed RBPF for 6-DOF localization in robotic peg-in-hole assembly with a static geometric map, bearing no connection to environmental fields.

Our results apply directly to underwater localization, which remains a major challenge. It often relies on costly sensing modalities that are not universally reliable, such as Doppler Velocity Loggers~\cite{teixeira2016auv}, acoustic positioning infrastructures, or acoustic and optical imaging systems that demand substantial onboard computation and that can fail due to lighting conditions and water turbidity conditions~\cite{mcconnell2022perception,islam2024computer}, and sonar-based sensors can suffer from low resolution.

Many spatiotemporal environmental phenomena of practical interest are governed by PDEs. The shallow water equation (SWE) describes water flows in coastal, estuarine, and riverine environments. The advection-diffusion equation describes the transport and mixing of scalar quantities such as temperature, salinity, dissolved oxygen, and chemical concentration in water bodies.  In each case, the PDE produces fields that naturally vary in space and time, and that can be predicted in future steps with current information. To address the challenges identified above, the main contributions of this paper are as follows:
\begin{itemize}
\item A water-feature-based localization framework that uses environmental fields governed by PDEs as localization signatures for GPS-denied operation. Two PDE classes are formulated: the shallow water equation and the advection-diffusion equation. 
\item An RBPF formulation that exploits the conditionally linear structure of the PDE measurement model. 
\item Simulation studies across two PDE scenarios: $i$) coastal wave propagation (SWE) and $ii$) water quality modeling (advection-diffusion), demonstrate that the RBPF consistently outperforms the standard PF.
\item Real-world validation using an autonomous surface vehicle measuring salinity, temperature, and dissolved oxygen, demonstrating that PDE-governed fields provide sufficient spatial variability for practical localization.
\end{itemize}

\section{Problem Formulation}

We consider a vehicle navigating a two-dimensional domain $\mathcal{W} \subseteq \mathbb{R}^2$ that is characterized by $M$ environmental scalar and vector fields governed by PDEs; these fields represent physical quantities such as water velocity, temperature, dissolved oxygen, or water height, and are collectively denoted by $\boldsymbol{\phi}:\mathcal{W}\times\mathbb{R}^+ \rightarrow \mathbb{R}^M$.

The vehicle state at time $t$ is $\mathbf{x}_t \in \mathcal{X}$, where $\mathcal{X}$ is the state space. The vehicle moves through $\mathcal{W}$ according to a kinematic motion model $\mathbf{f} :\mathcal{X}\times \mathcal{U} \rightarrow\mathcal{X}$, driven by a control input $\mathbf{u}_t$ from the set of admissible controls $\mathcal{U}$:
\begin{equation}
\label{eq:motion_model}
\mathbf{x}_{t+1} = \mathbf{f}(\mathbf{x}_t, \mathbf{u}_t).
\end{equation}
The projection $\Pi:\mathcal{X}\longrightarrow \mathcal{W}$ maps the full state to the vehicle position $\mathbf{p} = \Pi(\mathbf{x})$ in the domain. Moreover, the vehicle is equipped with $M$ onboard sensors, each measuring a corresponding channel of the environmental field $\boldsymbol{\phi}$. The observation model $\mathbf{h}: \mathcal{X} \longrightarrow \mathcal{Y}$ maps the vehicle state to the observation space $\mathcal{Y}$ as
\begin{equation}\label{eq:general-obs}
   \mathbf{h}(\mathbf{x}_t) =  \boldsymbol{\phi}(\Pi(\mathbf{x}_t),t) + \boldsymbol{\varepsilon}_t, \qquad \boldsymbol{\varepsilon}_t \sim \mathcal{N}(\mathbf{0}, \Sigma),
\end{equation}
where $\boldsymbol{\varepsilon}_t$ is zero-mean Gaussian measurement noise. Each observation therefore, provides a multimodal measurement vector that simultaneously captures multiple PDE-governed field quantities at the vehicle position.

\begin{problem}
  \textit{Given the motion model~\eqref{eq:motion_model}, the multimodal observation model~\eqref{eq:general-obs}, and the PDE-predicted environmental fields $\boldsymbol{\phi}$, the objective is to recursively estimate the vehicle state $\mathbf{x}_t$ at each time step $t$, conditioned on the history of observations $\mathbf{y}_{1:t}$ and the sequence of control inputs $\mathbf{u}_{1:t-1}$. In a nutshell, how to approximate the posterior distribution~$p(\mathbf{x}_t \mid \mathbf{y}_{1:t}, \mathbf{u}_{1:t-1})$ by exploiting the spatiotemporal structure of $\boldsymbol{\phi}$ as a dynamic localization reference? } 
\end{problem}

\section{Modeling Environmental Phenomena Utilizing PDEs}\label{sec:pde}

The solution of PDEs yields a set of spatially varying field quantities $\boldsymbol{\phi}=(\phi_1(\mathbf{x}, t), \ldots, \phi_M(\mathbf{x}, t))^\top$ in our domain of interest $\mathcal{W}$. The vehicle is equipped with $M$ onboard sensors that measure the local values of these fields at its current position $\mathbf{x} \in \mathcal{W}$. Because these fields exhibit distinct spatial variation, the resulting sensor readings provide sufficient information to constrain the vehicle's position, facilitating localization without the necessity of GPS.

This formulation remains general; while different PDE classes model various physical phenomena, the underlying localization structure is consistent. We focus on two specific instances relevant to maritime and coastal environments since they embody the behavior of most water features.

\subsubsection{Shallow Water Equation (SWE)}
\label{subsubsec:swe}

The shallow water equations model free-surface flows in coastal and riverine environments where the horizontal scale is much larger than the depth; it presents a useful framework for coastal or river tidal modeling. This model is an abstraction of the Navier-Stokes equations for fluid dynamics. It assumes that, at a given spatial coordinate $\mathbf{x}$, the column of water is moving as a single entity by averaging the velocities along it. This model represents conserved variables: water mass $\rho \eta:\mathcal{W}\times \mathbb{R}^+\rightarrow \mathbb{R}^+$, and its momentum $\rho\eta \mathbf{v}$. Here, $\rho$ is the fluid's density, $\mathbf{v}:\mathcal{W}\times \mathbb{R}^+\rightarrow \mathbb{R}^2$ groups the velocities in the $x$ and $y$ directions, respectively.
The SWE, in its conservative form, can be expressed in its block notation:

\begin{equation}
    \label{eq:shallow-water-equation}
    \begin{aligned}
        \frac{\partial \boldsymbol{\phi}}{\partial t} + \nabla \cdot 
        F(\boldsymbol{\phi})
         &= \mathbf{s}(\mathbf{x}, t,\boldsymbol{\phi}), \text{ for } (\mathbf{x}, t) \in \mathcal{W} \times \mathbb{R} \\
         F(\boldsymbol{\phi}) &= \eta \begin{bmatrix}
             \mathbf{v}^\top \\
             \mathbf{v} \otimes \mathbf{v} + \tfrac{1}{2}g\eta I 
         \end{bmatrix} \\
         \boldsymbol{\phi} &=\boldsymbol{\phi}_0, \text{ for } t = 0,
    \end{aligned}
\end{equation}

where $\otimes$ is the exterior product between two vectors, $g$ represents gravitational acceleration, $I$ is the $2\times2$ identity matrix, $\boldsymbol{\phi}_0$ is the initial condition encompassing initial water height and velocity, and $\mathbf{s}$ denotes the forcing term increasing or decreasing any variable. It is worth noting that even though $\mathbf{s}$ still depends on the conserved variables, it is considered a forcing term due to its nature of incorporating external information into the model. Furthermore, we purpusely included the variables that $\mathbf{s}$ depend on, this will facilitate introducing typical forcing tems that are presented in Section \ref{sec:results}. On the other hand, if we assume the fluid is incompressible -as water is close to being- its density can be removed from the equation, simplifying it. In this context, the vehicle measurement vector is $\boldsymbol{\phi} = (\eta,u, v)^\top$, corresponding to $M = 3$ sensing channels.

\subsubsection{Advection-Diffusion Equation}
\label{subsubsec:advdiff}

To model the transport of scalar water properties such as temperature, salinity, chlorophyll, or chemical concentrations, we utilize the advection-diffusion equation; it balances the behavior when the main evolution of such properties is determined by transportation and dissolution factors. For a scalar field $\phi(\mathbf{x}, t)$, the dynamics are governed by:

\begin{equation}
\label{eq: advection-diffusion}
    \begin{aligned}
    \frac{\partial \phi}{\partial t} + \nabla \cdot( \phi \mathbf{v} - \kappa \nabla \phi) &= s, \text{ for } (\mathbf{x}, t) \in \mathcal{W} \times \mathbb{R} \\
    \phi &=\phi_0, \text{ for } t = 0,      
    \end{aligned}
\end{equation}

where $\mathbf{v}$ is the water velocity field, $\kappa$ is the diffusivity term, and $s$ represents the source term. The diffusion term $-\kappa \nabla \phi$ accounts for the diffusion of the property from regions of high concentration to the regions with low concentration, while the advection term $\phi \mathbf{v}$ describes its transport by the fluid flow.

In oceanographic applications, distinct scalar fields $\phi_i$, $i=1,\ldots, M$, that satisfy equation \eqref{eq: advection-diffusion} may be coupled in the same region of interest with specific coefficients and source terms by the same velocity field. Similar to the SWE, by vectorizing $\boldsymbol{\phi} = (\phi_1,\ldots,\phi_M)^\top$, $\boldsymbol{s} = (s_1,\ldots,s_M)^\top$, $\boldsymbol{\phi}_0 = ({\phi_0}_1,\ldots,{\phi_0}_M)^\top$, $\boldsymbol{\kappa} = (\kappa_1,\ldots,\kappa_M)^\top$ and defining $\diag(\mathbf{z})$ as the diagonal matrix whose entrances are filled with the elements of $\mathbf{z}$; the different features can be written as a coupled PDE as 

\begin{equation}
\label{eq: advection-diffusion-coupled}
    \begin{aligned}
    \frac{\partial \boldsymbol{\phi}}{\partial t} + \nabla \cdot(  \boldsymbol{\phi} \otimes \mathbf{v} - \diag(\boldsymbol{\kappa}) \nabla \boldsymbol{\phi}) &= \mathbf{s}, \text{ for } (\mathbf{x}, t) \in \mathcal{W} \times \mathbb{R} \\
    \boldsymbol{\phi} &=\boldsymbol{\phi}_0, \text{ for } t = 0.         
    \end{aligned}
\end{equation}

In this case, the vehicle measurement vector is $\boldsymbol{\phi}$ containing $M$ channels.

\section{Localization with Rao-Blackwellized Particle Filter (RBPF)}
\label{sec:particle-filter}

\subsection{State Partition}
Unlike standard particle filter which samples the full state and computes importance weights from the measurement likelihood, the Rao-Blackwellized particle filter (RBPF) exploits posterior factorization when the state admits a conditionally linear substructure, resulting in reducing the variance of the posterior estimate while requiring fewer particles~\cite{doucet2000sequential, murphy2001rao} as:
\begin{equation}\label{eq:rb-factorization}
    p(\mathbf{x}_t^n, \mathbf{x}_t^l \mid \mathbf{y}_{1:t}) = p(\mathbf{x}_t^n \mid \mathbf{y}_{1:t})\, p(\mathbf{x}_t^l \mid \mathbf{x}_t^n, \mathbf{y}_{1:t}),
\end{equation}
where the nonlinear states $\mathbf{x}_t^n$ are represented by particles, and the conditionally linear states $\mathbf{x}_t^l$ are tracked analytically via a Kalman filter conditioned on each particle~\cite{schon2005marginalized}.

In our setup, we partition the full state into a nonlinear component $\mathbf{x}_t^n$ and a linear component $\mathbf{x}_t^l$. The nonlinear state captures the vehicle pose and speed as: $  \mathbf{x}_t^n = \left[ p_x,  p_y, \theta,  v \right]^\top \in \mathbb{R}^4$,
where $(p_x, p_y)$ is the position, $\theta$ is the heading, and $v$ is the forward speed. The linear state models a slowly varying sensor bias on each measurement channel as: $   \mathbf{x}_t^l = \left[ \delta\boldsymbol{\phi}_1,  \cdots, \delta\boldsymbol{\phi}_M \right]^\top \in \mathbb{R}^M.$
With this decomposition, the measurement model becomes conditionally linear:
\begin{equation}\label{eq:meas-model}
    \mathbf{y}_t = \mathbf{h}(\mathbf{x}_t^n) + H\, \mathbf{x}_t^l,
\end{equation}
remembering that $\mathbf{h}$ is the nominal PDE evaluated at the particle position (see equation~\eqref{eq:general-obs}) and $H = I_M$ is the identity observation matrix for the bias states. Conditioned on a given particle position $\mathbf{x}_t^{n,(i)}$, equation~\eqref{eq:meas-model} is linear in $\mathbf{x}_t^l$, which is exactly the structure required for Rao-Blackwellization.

\subsection{Motion Model}

The RBPF maintains $N$ particles. Each particle $i$ is represented by the tuple
$\{\mathbf{x}_t^{n,(i)}, \hat{\mathbf{x}}_t^{l,(i)}, P_t^{(i)}, w_t^{(i)}\}$, where
$\mathbf{x}_t^{n,(i)}$ denotes the sampled nonlinear state,
$\hat{\mathbf{x}}_t^{l,(i)}$ and $P_t^{(i)}$ are the conditional Kalman filter mean and covariance of the linear state, and
$w_t^{(i)}$ is the associated importance weight.
The nonlinear component evolves according to a unicycle-type motion model driven by the control input
$\mathbf{u}_t = (v_{\text{cmd}}, \omega_{\text{cmd}})^\top$:
\begin{equation}\label{eq:motion-model}
    \mathbf{x}_{t+1}^n = \mathbf{f}(\mathbf{x}_t^n, \mathbf{u}_t) + \boldsymbol{\omega}_t,
    \qquad \boldsymbol{\omega}_t \sim \mathcal{N}(\mathbf{0}, \Sigma_{\mathbf{f}}),
\end{equation}
where $\Sigma_{\mathbf{f}}$ is the covariance of the additive process noise affecting the nonlinear state.
The linear bias state evolves according to an identity state-transition model driven by zero-mean Gaussian white noise:
\begin{equation}\label{eq:lin-process}
    \mathbf{x}_{t+1}^l = \mathbf{x}_t^l + \boldsymbol{\zeta}_t,
    \qquad \boldsymbol{\zeta}_t \sim \mathcal{N}(\mathbf{0}, \Lambda),
\end{equation}
where $\Lambda$ is a diagonal covariance matrix modeling slow-varying sensor drift.

\subsection{Measurement Update}

At each time step, the RBPF incorporates a new observation $\mathbf{y}_t \in \mathbb{R}^M$
of the $M$ PDE field channels, as defined in~\eqref{eq:general-obs}. The update
performs two coupled operations for each particle $i$, i.e., computing the importance
weight and updating the Kalman filter for the linear substate $\mathbf{x}_t^l$.

The linear substate $\mathbf{x}_t^l \in \mathbb{R}^M$
represents a slowly varying additive sensor bias on each field channel. The
measurement model at particle $i$ is
\begin{equation}\label{eq:meas-model}
    \mathbf{y}_t = \boldsymbol{\phi}\!\left(\mathbf{x}_t^{n,(i)}, t\right) + H\, \mathbf{x}_t^{l,(i)} + \boldsymbol{\varepsilon}_t,
    \qquad \boldsymbol{\varepsilon}_t \sim \mathcal{N}(\mathbf{0}, \Sigma),
\end{equation}

\subsubsection*{Innovation}

For each particle $i$, the nominal field vector $\boldsymbol{\phi}(\mathbf{x}_t^{n,(i)}, t)$ is obtained
by rounding the particle's continuous position to the nearest grid index and reading
off the stored field values. If the cell lies outside the domain boundary or is a
dry (masked) cell, the particle receives zero weight. Otherwise, the innovation is
\begin{equation}\label{eq:innovation}
    \mathbf{e}_t^{(i)} = \mathbf{y}_t - \boldsymbol{\phi}\!\left(\mathbf{x}_t^{n,(i)}, t\right) - H\, \hat{\mathbf{x}}_{t|t-1}^{l,(i)},
\end{equation}
and the innovation covariance is
\begin{equation}\label{eq:innov-cov}
    S_t^{(i)} = H\, P_{t|t-1}^{(i)}\, H^\top + \Sigma.
\end{equation}

\subsubsection*{Weight Update}

The importance weight is proportional to the marginal likelihood of $\mathbf{y}_t$ given
the particle's nonlinear state, obtained by integrating out $\mathbf{x}_t^l$ analytically.
This marginal likelihood is Gaussian because the measurement model is conditionally
linear in $\mathbf{x}_t^l$, which is the core benefit of Rao-Blackwellization. The log-weight
is computed via the Cholesky factorization $S_t^{(i)} = L_t^{(i)} {L_t^{(i)}}^\top$
for numerical stability:
\begin{equation}\label{eq:log-weight}
\begin{split}
    \log \tilde{w}_t^{(i)} &= - \sum_{k=1}^M \log \bigl[L_t^{(i)}\bigr]_{kk} \\
    &\quad - \frac{1}{2} \left\| {L_t^{(i)}}^{-1} \mathbf{e}_t^{(i)} \right\|^2
    - \frac{M}{2} \log 2\pi,
\end{split}
\end{equation}
which corresponds to evaluating $\log \mathcal{N}(\mathbf{e}_t^{(i)};\, \mathbf{0},\, S_t^{(i)})$.
Particles in dry or out-of-bounds cells are assigned $\log \tilde{w}_t^{(i)} =
-\infty$, enforcing the physical constraint that the vehicle must remain in
navigable water. Weights are then normalized using the log-sum-exp identity:
\begin{equation}\label{eq:weight-normalize}
    w_t^{(i)} =
    \frac{%
        \exp\!\left(\log \tilde{w}_t^{(i)} - c\right)
    }{%
        \displaystyle\sum_{j=1}^{N} \exp\!\left(\log \tilde{w}_t^{(j)} - c\right)
    },
    \qquad c = \max_{1\leq j\leq N}\, \log \tilde{w}_t^{(j)}.
\end{equation}

\subsubsection*{Kalman Filter Update}

The Kalman filter for particle $i$ is updated via the standard equations:
\begin{align}
    K_t^{(i)} &= P_{t|t-1}^{(i)}\, H^\top \left(S_t^{(i)}\right)^{-1},
    \label{eq:kalman-gain} \\
    \hat{\mathbf{x}}_{t|t}^{l,(i)} &= \hat{\mathbf{x}}_{t|t-1}^{l,(i)} + K_t^{(i)}\, \mathbf{e}_t^{(i)},
    \label{eq:kf-mean-update} \\
    P_{t|t}^{(i)} &= P_{t|t-1}^{(i)} - K_t^{(i)}\, S_t^{(i)}\, {K_t^{(i)}}^\top.
    \label{eq:kf-cov-update}
\end{align}
Each particle carries its own Kalman filter that tracks the per-channel sensor
bias conditioned on that particle's trajectory hypothesis.

\subsection{Prediction and Resampling}

The RBPF prediction step differs from a standard particle filter by performing
ancestor resampling before propagating the nonlinear state. At the start of each
time step, ancestor indices are drawn via systematic resampling from the current
weight distribution. Each particle $i$ then inherits the nonlinear state, bias
mean, and bias covariance of its chosen ancestor, and all weights are reset to
uniform. The nonlinear state is then propagated by the motion model with additive
noise:
\begin{equation}\label{eq:motion-model}
    \mathbf{x}_t^{n,(i)} = \mathbf{f}\!\left(\mathbf{x}_{t-1}^{n,(a_i)},\; \mathbf{u}_t + \boldsymbol{\omega}_t^{(i)}\right),
    \qquad \boldsymbol{\omega}_t^{(i)} \sim \mathcal{N}(\mathbf{0},\, \Sigma_{\mathbf{f}}),
\end{equation}
where $a_i$ is the ancestor index for particle $i$ and $\Sigma_{\mathbf{f}}$ is the motion
noise covariance acting on the control inputs. The linear state covariance is
updated by adding the bias drift noise:
\begin{equation}\label{eq:kf-predict}
    \hat{\mathbf{x}}_{t|t-1}^{l,(i)} = \hat{\mathbf{x}}_{t-1}^{l,(a_i)},
    \qquad
    P_{t|t-1}^{(i)} = P_{t-1}^{(a_i)} + \Lambda,
\end{equation}
where $\Lambda$ is the bias process noise covariance (c.f. equation~\eqref{eq:lin-process}). The bias mean is unchanged in the prediction step since the bias is modeled as a random walk.

After the measurement update in~\eqref{eq:weight-normalize}, particle degeneracy
is monitored via the effective sample size
\begin{equation}\label{eq:neff}
    N_{\mathrm{eff}} = \Bigg(\sum_{i=1}^{N} \left(w_t^{(i)}\right)^2\Bigg)^{-1}.
\end{equation}
A secondary SIR resampling step is triggered when $N_{\mathrm{eff}} < N/2$.
During this step, the nonlinear state $\mathbf{x}_t^{n,(i)}$ and the associated Kalman
filter $(\hat{\mathbf{x}}_t^{l,(i)}, P_t^{(i)})$ are duplicated together, preserving
the per-particle posterior on the linear substate. After resampling, all weights
are reset to $1/N$. The complete RBPF cycle is summarized in
Algorithm~\ref{alg:rbpf}.

\begin{algorithm}[t]
\caption{RBPF for PDE-based Vehicle Localization}\label{alg:rbpf}
\begin{algorithmic}[1]
\Require Particle set
    $\bigl\{\mathbf{x}_{t-1}^{n,(i)},\, \hat{\mathbf{x}}_{t-1}^{l,(i)},\, P_{t-1}^{(i)},\,
    w_{t-1}^{(i)}\bigr\}_{i=1}^{N}$, control $\mathbf{u}_t$, observation
    $\mathbf{y}_t \in \mathbb{R}^M$
\Ensure Updated particle set
    $\bigl\{\mathbf{x}_t^{n,(i)},\, \hat{\mathbf{x}}_t^{l,(i)},\, P_t^{(i)},\,
    w_t^{(i)}\bigr\}_{i=1}^{N}$
\State \textbf{Ancestor resampling:} Draw indices $\{a_i\}$ by systematic
    resampling from $\{w_{t-1}^{(i)}\}$; reset $w^{(i)} \gets 1/N$
\For{$i = 1, \ldots, N$}
    \State \textbf{Predict nonlinear:}
        $\mathbf{x}_t^{n,(i)} \gets \mathbf{f}\!\left(\mathbf{x}_{t-1}^{n,(a_i)},\, \mathbf{u}_t + \boldsymbol{\omega}_t^{(i)}\right)$,
        \quad $\boldsymbol{\omega}_t^{(i)} \sim \mathcal{N}(\mathbf{0}, \Sigma_{\mathbf{f}})$
    \State \textbf{Predict linear:}
        $\hat{\mathbf{x}}_{t|t-1}^{l,(i)} \gets \hat{\mathbf{x}}_{t-1}^{l,(a_i)}$,\quad
        $P_{t|t-1}^{(i)} \gets P_{t-1}^{(a_i)} + \Lambda$
    \State \textbf{Innovation:}
        $\mathbf{e}_t^{(i)} \gets \mathbf{y}_t - \boldsymbol{\phi}(\mathbf{x}_t^{n,(i)}, t) - H\,\hat{\mathbf{x}}_{t|t-1}^{l,(i)}$
    \State \textbf{Innovation covariance:}
        $S_t^{(i)} \gets H P_{t|t-1}^{(i)} H^\top + R$;\quad
        factor $S_t^{(i)} = L_t^{(i)}{L_t^{(i)}}^\top$ via Cholesky
    \State \textbf{Log-weight:}
        $\log \tilde{w}_t^{(i)} \gets \log w_{t-1}^{(i)} +
        \log \mathcal{N}\!\left(\mathbf{e}_t^{(i)};\, \mathbf{0},\, S_t^{(i)}\right)$
        via~\eqref{eq:log-weight}
    \State \textbf{Kalman update:} Compute $K_t^{(i)}$; update
        $\hat{\mathbf{x}}_{t|t}^{l,(i)}$, $P_{t|t}^{(i)}$
        via~\eqref{eq:kalman-gain}--\eqref{eq:kf-cov-update}
\EndFor
\State \textbf{Normalize:} $w_t^{(i)} \gets \exp(\log \tilde{w}_t^{(i)} - c)\;/
    \sum_j \exp(\log \tilde{w}_t^{(j)} - c)$ via~\eqref{eq:weight-normalize}
\If{$N_{\mathrm{eff}} < N/2$}
    \State SIR resample particles with associated
        $(\hat{\mathbf{x}}_t^{l,(i)}, P_t^{(i)})$; reset weights to $1/N$
\EndIf
\end{algorithmic}
\end{algorithm}

\section{Results}\label{sec:results}

This section aims to showcase two case studies in different areas using the two presented PDE models, given that they require different types of data to be simulated and validated. All experiments and simulations are conducted using a general-purpose laptop computer. For simulations, the measurement and motion noise covariances are set to $\mathrm{diag}(0.2^2, 0.2^2, 0.1^2)$ and $\mathrm{diag}(0.5^2, 0.087^2)$, respectively. The linear process noise is set to $\mathrm{diag}(10^{-8}, 10^{-8}, 10^{-8})$.

\subsection{Case Study 1: Shallow Water Equations}\label{subsec:swe}

In this scenario, we target the study of equations that arise as the solutions of \eqref{eq:shallow-water-equation}, the robot has access to $M=3$ measurement channels, that is, the water height $\eta$ and the water velocities $\mathbf{v}= (u,v)$. As mentioned before, we made explicit the dependence of the source term $\mathbf{s}$ on $\boldsymbol{\phi}$. To have realistic conditions from small- to medium-sized coastal areas, we incorporate two terms to $\mathbf{s}$: bathymetry $b:\mathcal{W}\rightarrow \mathbb{R}$ and wind speed $\boldsymbol{\omega}:\mathcal{W}\times\mathbb{R}^+\rightarrow \mathbb{R}^2$. Consicely, redefine 
\begin{equation}
    \label{eq:swe-forcing-term}
    \begin{aligned}
        \mathbf{s} &= \begin{bmatrix}
            0 \\
            \boldsymbol{\tau}(\mathbf{v}-\boldsymbol{\omega}) -g \cdot(\nabla b)^\top
        \end{bmatrix},\\
        \boldsymbol{\tau}(\mathbf{z})&=\frac{\rho_{\text{water}}}{\rho_{\text{air}}}c_{\text{c}}\|\mathbf{z}\|\mathbf{z}.
    \end{aligned}
\end{equation}

$\boldsymbol{\tau}$ encompasses the wind stress tensor, $\rho_{\text{water}}$, $\rho_{\text{air}}$, and $c_{\text{c}}$ are the water density, the air density, and the drag coefficient, respectively. The data used to estimate the bathymetry were obtained from the General Bathymetric Chart of the Oceans (GEBCO), and a region of interest was selected around a designated coastal area. On the other hand, we obtained the wind velocities from recorded conditions around this area when the wind has a significant impact on the water's velocity $\|\boldsymbol{\omega} \| \approx 12 \; \text{m}/\text{s}$.
\definecolor{viriMin}{HTML}{440154}
\definecolor{viriMax}{HTML}{fde725}
\begin{figure}[htbp]
    \centering
    \begin{subfigure}[b]{0.49\linewidth}
        \centering
        \includegraphics[width=\linewidth]{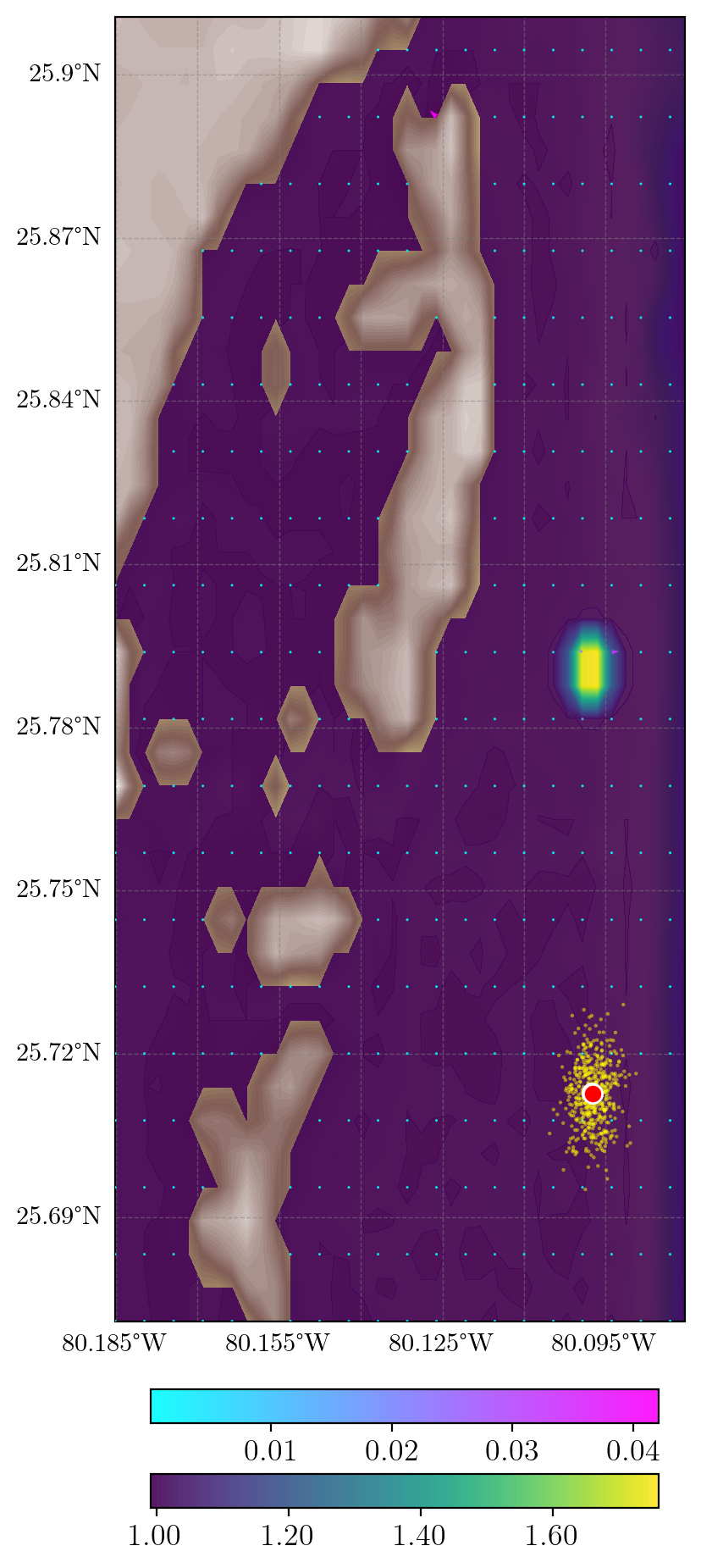}
        \caption{$t = 0$ s}
        \label{fig:swe_sim_t2}
    \end{subfigure}
    \begin{subfigure}[b]{0.49\linewidth}
        \centering
        \includegraphics[width=\linewidth]{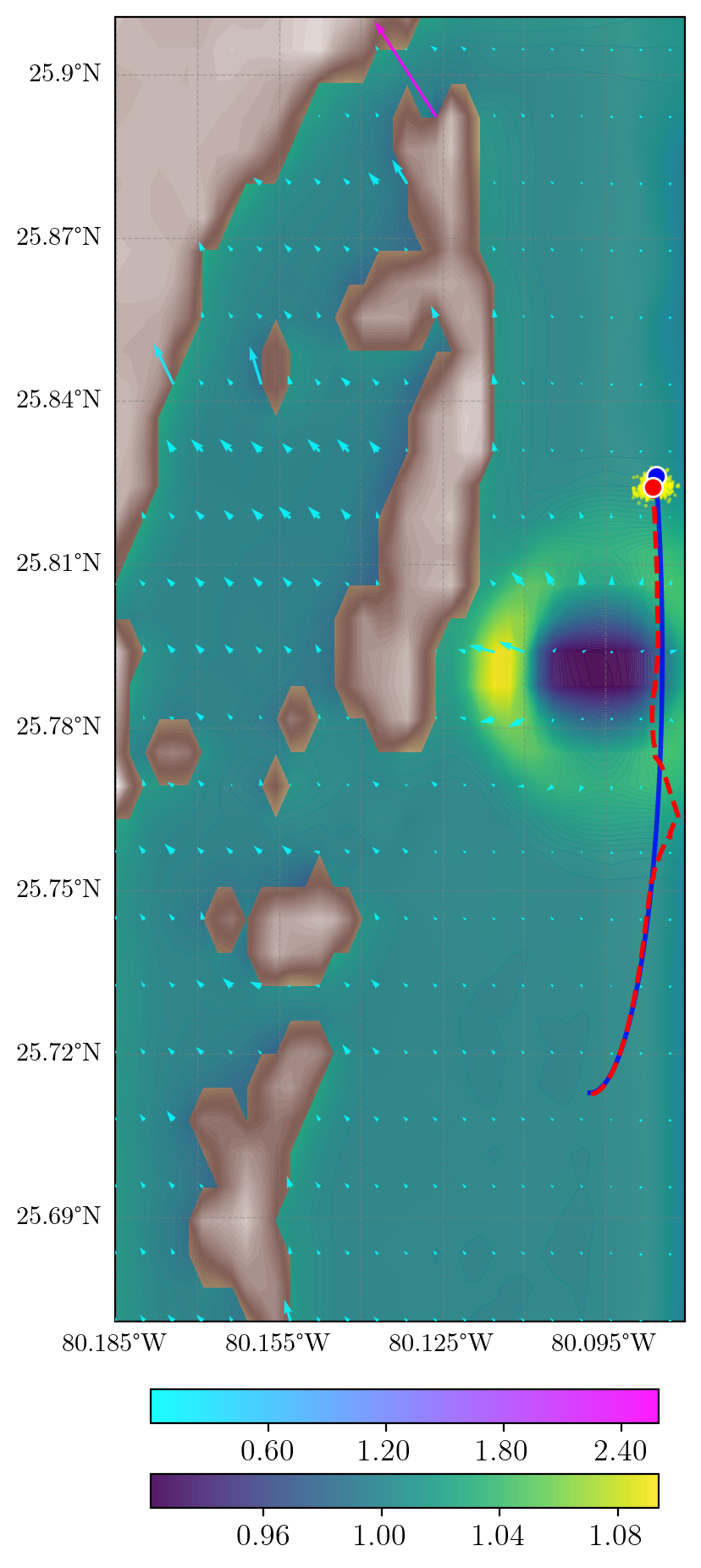}
        \caption{$t = 160$ s}
        \label{fig:swe_sim_t3}
    \end{subfigure}
    \begin{subfigure}[b]{0.49\linewidth}
        \centering
        \includegraphics[width=\linewidth]{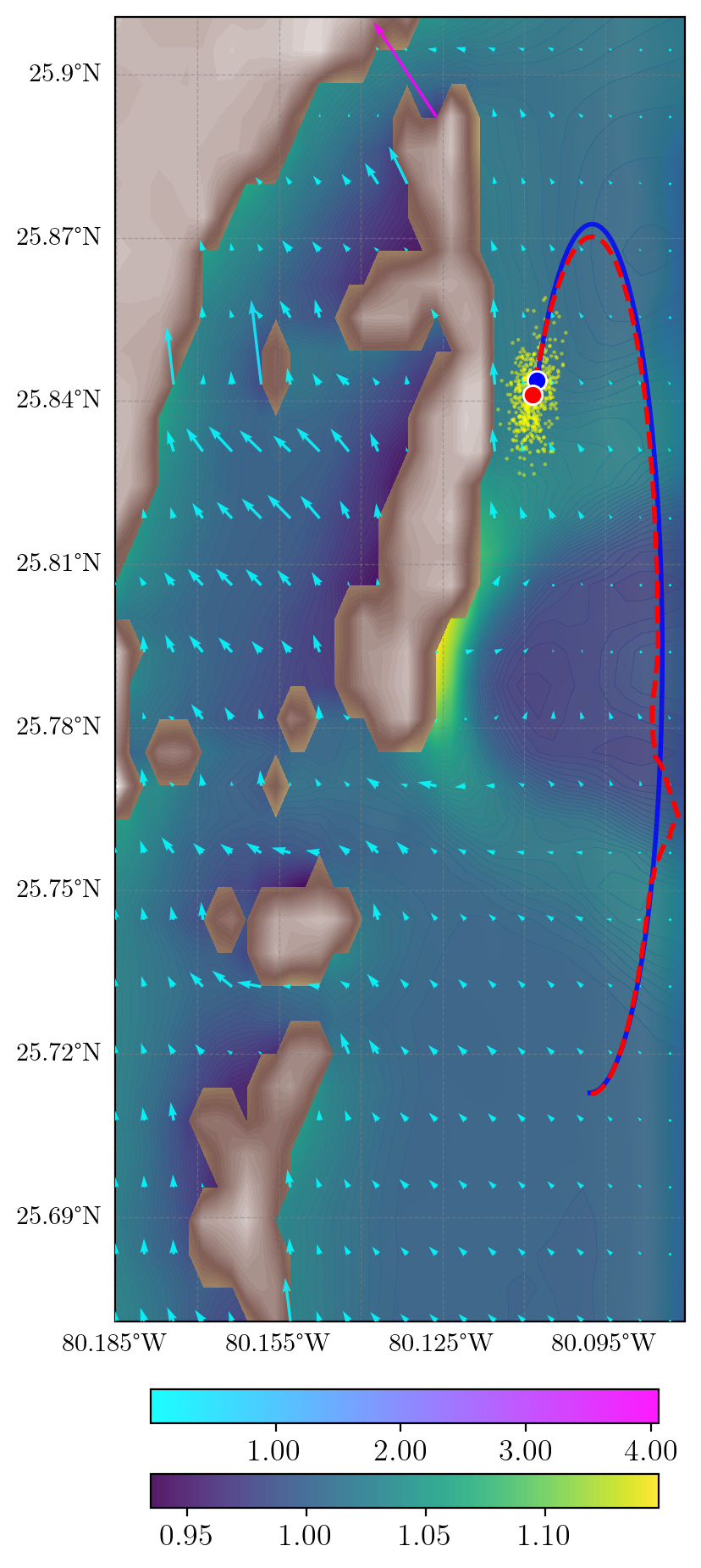}
        \caption{$t = 330$ s}
        \label{fig:swe_sim_t6}
    \end{subfigure}
    \begin{subfigure}[b]{0.49\linewidth}
        \centering
        \includegraphics[width=\linewidth]{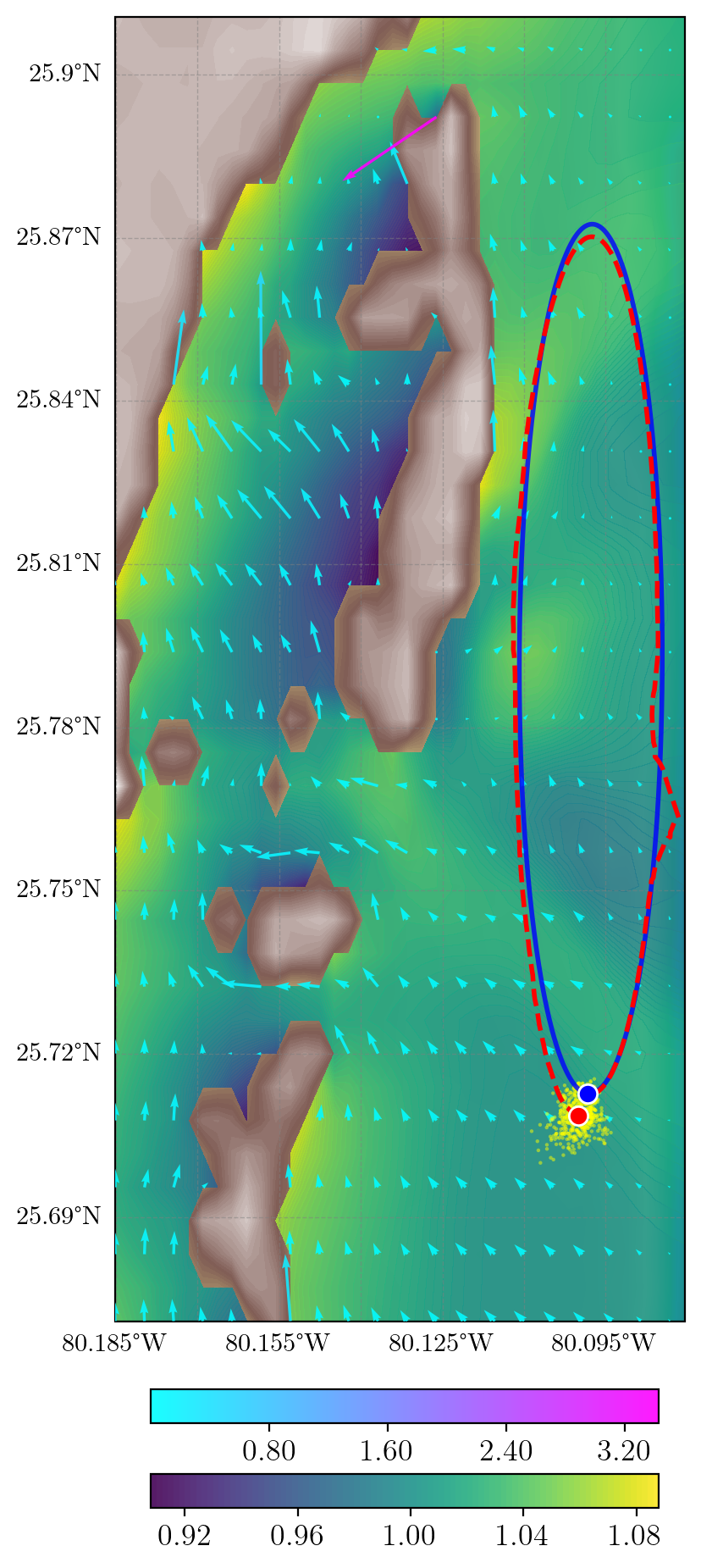}
        \caption{$t = 500$ s}
        \label{fig:swe_sim_t11}
    \end{subfigure}
    \caption{RBPF localization at different timestamps showing the estimation of the robot state over time, solving the SWE~\eqref{eq:shallow-water-equation}. The SWE incorporates the \textcolor{viriMin}{water height} (m) corrected by the bathymetry (i.e., water above sea level); the \textcolor{cyan}{velocity field} (ms$^{-1}$) is plotted as a vector field; and the \textcolor{brown}{terrain above sea level} defines the coastal domain. The robot is localizing itself by reconstructing an \textcolor{red}{estimated trajectory} from an elliptical \textcolor{blue}{ground truth trajectory} using 500 RBPF \textcolor{Goldenrod}{particles}.}
    \label{fig:swe_localization_sim}
\end{figure}

Figure~\ref{fig:swe_localization_sim} shows multiple snapshots of this model simulating how a water wave propagates towards the coast; in particular, it displays the water height from the surface, defined as the water height corrected by the bathymetry $\eta - b$. Moreover, it shows the robot traversing an elliptical path (ground truth) while using the RBPF to localize itself by measuring the water velocities and water height $\boldsymbol{\phi} = (\eta, u, v)$. The RBPF uses $500$ particles, simulation step time is set to $1\,\text{s}$, and a total simulation time of $500\,\text{s}$. As can be observed in Figure~\ref{fig:swe_localization_sim}, the particles start scattered. In the absence of field measurements, the estimated red trajectory deviates from the desired path. However, when the vehicle encounters the field's noisy observations, the particles quickly converge toward the true vehicle state.

\subsection{Case Study 2: Advection-Diffusion}\label{subsec:adv-diff} 

The other motivating scenario includes the incorporation of other key features that are of major importance in multiple environmental surveillance, monitoring, and restoration efforts. We bring again equation~\eqref{eq: advection-diffusion-coupled}, which shows that we need to define the diffusivity constants $\boldsymbol{\kappa}$, which depend on each water feature and the velocity field $\mathbf{v}$ that couples all the equations together. The diffusivity constants vary between $0.01 \text{m}^2/\text{s}$ and $1\text{m}^2/\text{s}$ and the water velocity field averages $\|\mathbf{v}\|=1.0\text{m}/\text{s}$. We utilized data previously acquired to fit the initial condition $\boldsymbol{\phi}_0$ of the PDE. 

\begin{figure}[th]
    \centering
    \begin{subfigure}[b]{\linewidth}
        \centering
        \includegraphics[width=\linewidth]{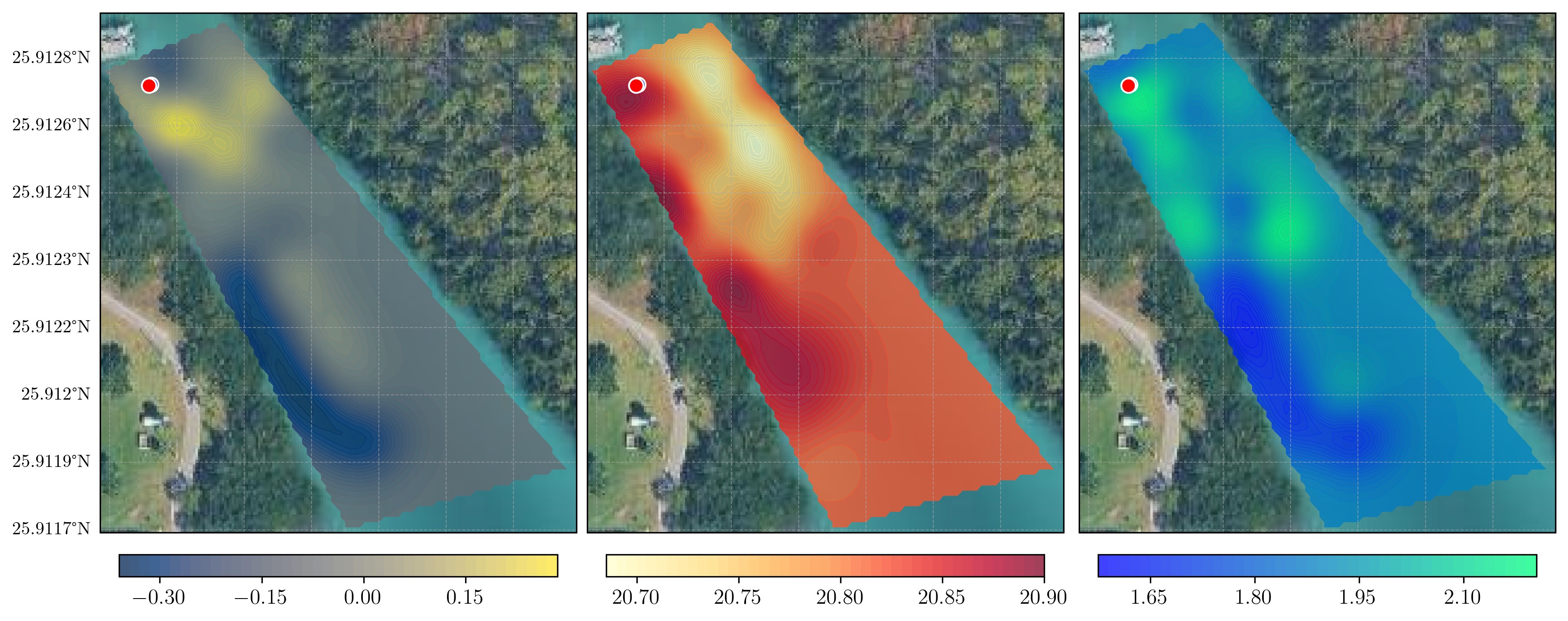}
        \caption{$t = 0$ s}
        \label{fig:adv-diff1}
    \end{subfigure}
    \begin{subfigure}[b]{\linewidth}
        \centering
        \includegraphics[width=\linewidth]{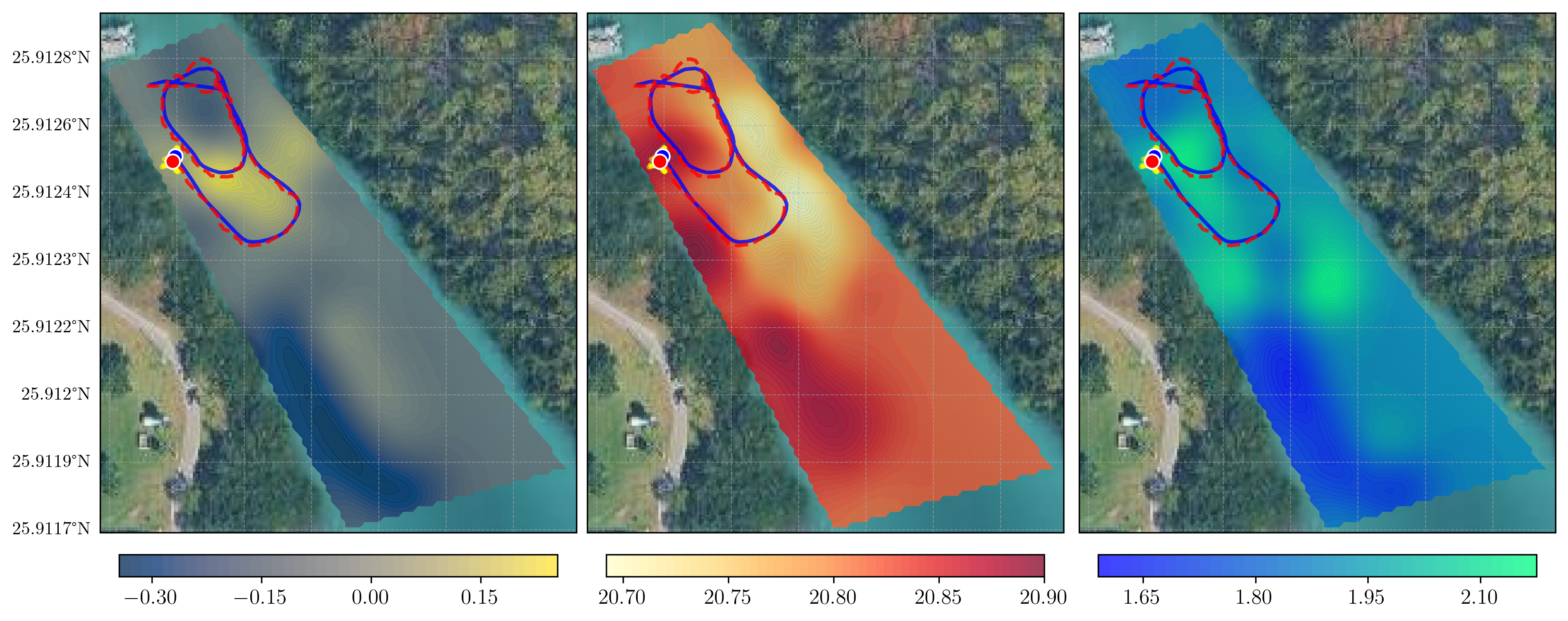}
        \caption{$t = 160$ s}
        \label{fig:adv-diff2}
    \end{subfigure}
    \begin{subfigure}[b]{\linewidth}
        \centering
        \includegraphics[width=\linewidth]{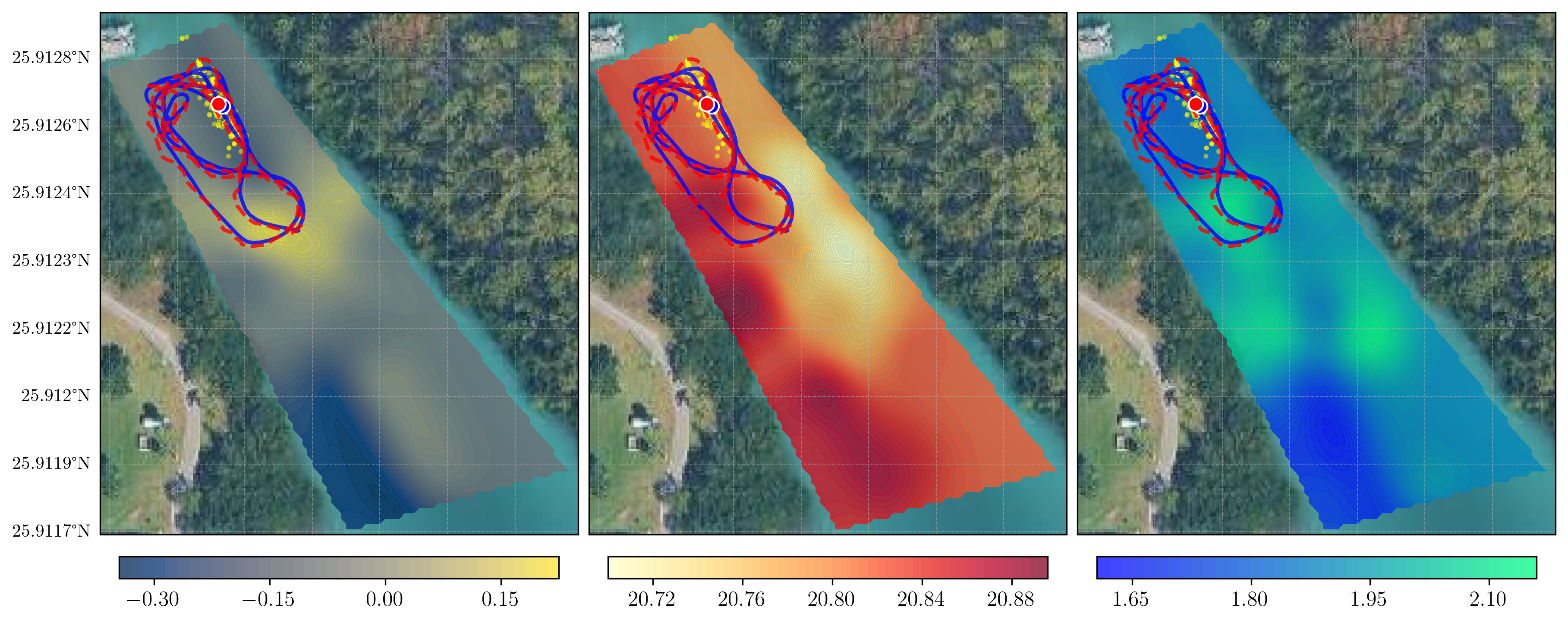}
        \caption{$t = 320$ s}
        \label{fig:adv-diff3}
    \end{subfigure}
    \begin{subfigure}[b]{\linewidth}
        \centering
        \includegraphics[width=\linewidth]{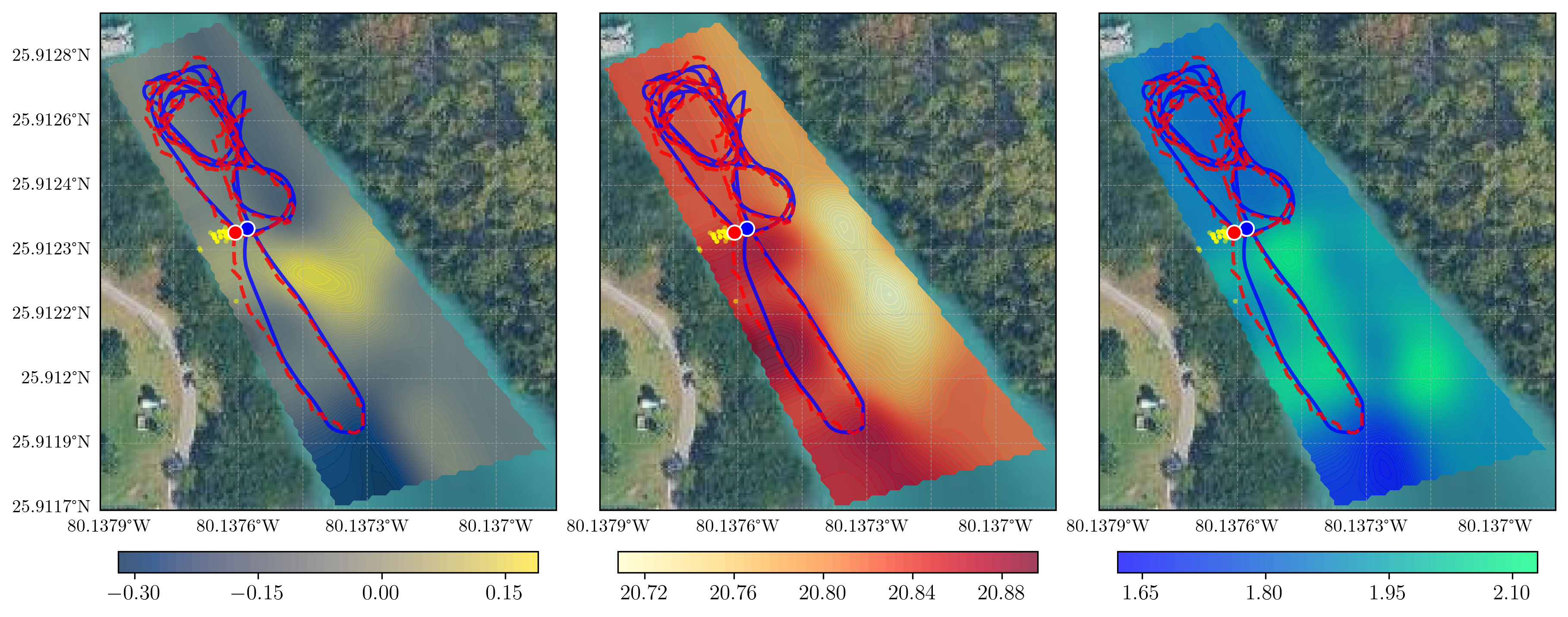}
        \caption{$t = 500$ s}
        \label{fig:adv-diff4}
    \end{subfigure}
    \caption{RBPF localization at different timestamps showing the estimation of the robot state over time. The robot estimates the \textcolor{blue}{ground truth path} by estimating its \textcolor{red}{current trajectory} using the \textcolor{Goldenrod}{RBPF} particles. The robot has access to $M=3$ fields from left to right: Turbidity (FNU), Temperature ($^\circ$C), and Chlorophyll ($\mu$gL$^{-1}$).}
    \label{fig:adv-diff-sim}
\end{figure}

Figure~\ref{fig:adv-diff-sim} shows multiple snapshots of this model simulating how environmental quantities evolve over a total simulation time of $500$\,s with data collected at $1$\,Hz. Here, we chose three water features: chlorophyll ($\mu$g/L), turbidity in Formazin Nephelometric Units (FNU), and temperature ($^\circ$C).
Similar to the SWE simulation, we use the RBPF with $500$ particles. The blue trajectories represent the ground truth collected by the GPS sensor, while the red trajectories are the RBPF estimates. As can be observed in Figure~\ref{fig:adv-diff-sim}, the RBPF accurately estimates the vehicle state when all three measurement channels are fused. In the absence of spatially distinctive measurements, the estimated trajectories deviate from the ground truth.

\subsection{Performance Analysis}\label{subsec:performance}

To evaluate filter performance, we report two complementary metrics: final error and root mean square error (RMSE). The final error captures the position error at the last time step, providing a direct measure of whether the filter has converged by the end of the run, though it reflects only a single snapshot. RMSE, on the other hand, aggregates squared errors across all time steps, providing a measure of average tracking accuracy over the entire trajectory. Reporting both together gives a more complete picture: a filter with low final error but high RMSE performed poorly early and recovered late, while one with low RMSE but high final error tracked well throughout but diverged near the end.

\pgfplotsset{
    myboxplotstyle/.style={
        boxplot/draw direction=y,
        boxplot={
            draw position={1/3 + floor(\plotnumofactualtype/2) + 1/3*mod(\plotnumofactualtype,2)},
            box extend=0.3,
        },
        x=2cm, %
        height = 4.5cm,
        xtick={0,1,2,3,4,5},
        x tick label as interval,
        xticklabels={%
            {100},
            {200},
            {300},
            {400},
            {500},
        },
        x tick label style={
            align=center
        },
        ymajorgrids=true,
        grid style=dashed,
        cycle list={
            {RoyalBlue, fill=RoyalBlue!20},
            {Orange, fill=Orange!20}
        },
    }
}

\begin{figure}[h!]
    \centering
    \begin{subfigure}[b]{\linewidth}
    \resizebox{\linewidth}{!}{
    \begin{tikzpicture}
    \begin{axis}[ylabel={RMSE (m)},
            myboxplotstyle,
            xticklabels={},
            ]
    \foreach \particle in {100, 200, 300, 400, 500}{
        \edef\colnameA{RMSE_m\string_0\string_\particle}
        \edef\colnameB{RMSE_m\string_1\string_\particle}
        
        \addplot+ [boxplot] 
            table[y=\colnameA, col sep=comma] {results/swe_benchmark_results.csv};
        \addplot+ [boxplot] 
            table[y=\colnameB, col sep=comma] {results/swe_benchmark_results.csv};
        }
        \end{axis}
        \end{tikzpicture}        }
    \end{subfigure}
    
    \begin{subfigure}[b]{\linewidth}
    \resizebox{\linewidth}{!}{
    \begin{tikzpicture}
    \begin{axis}[ylabel={Final Position Error (m)},
            myboxplotstyle
            ]
    
    \foreach \particle in {100, 200, 300, 400, 500}{
        \edef\colnameA{Final_Error\string_0\string_\particle}
        \edef\colnameB{Final_Error\string_1\string_\particle}
        
        \addplot+ [boxplot] 
            table[y=\colnameA, col sep=comma] {results/swe_benchmark_results.csv};
        \addplot+ [boxplot] 
            table[y=\colnameB, col sep=comma] {results/swe_benchmark_results.csv};
        }
        
        \end{axis}
        \end{tikzpicture}
        }
        \end{subfigure}
        
        \caption{Comparison of \textcolor{RoyalBlue}{PF} and \textcolor{Orange}{RBPF} localisation accuracy on the
                 shallow-water estimation (SWE) task across five independent
                 runs for varying numbers of particles.
                 Both metrics decrease as the particle count increases, with
                 RBPF consistently outperforms PF at equivalent sample sizes due to its Rao-Blackwellised marginalisation of the linear
                 sub-structure.}
        \label{fig:pf_rbpf_comparison}
\end{figure}
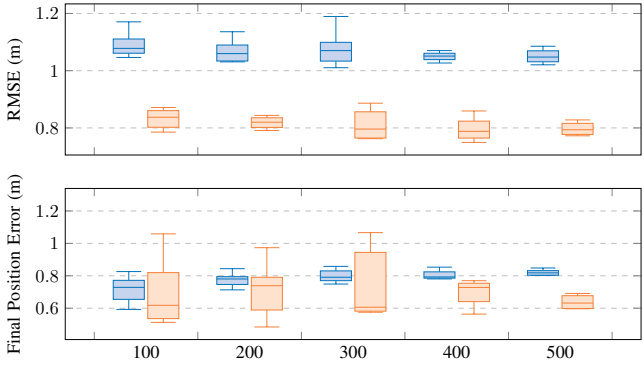

For the SWE model in Figure~\ref{fig:pf_rbpf_comparison}, the RBPF consistently outperforms the standard PF in sustained tracking accuracy. At 500 particles, the RBPF achieves an RMSE of $0.80 \pm 0.03$\,m compared to the PF's $1.07 \pm 0.04$\,m—a roughly 25\% reduction—while also attaining a lower final error ($0.70 \pm 0.15$\,m vs.\ $0.82 \pm 0.02$\,m). Notably, the RBPF with only 100 particles ($\text{RMSE} = 0.84 \pm 0.04$\,m) already surpasses the PF at 500 particles, demonstrating substantially greater sample efficiency. Across all particle counts, the PF's RMSE plateaus near $1.07$--$1.10$\,m, whereas the RBPF steadily improves from $0.84$\,m to $0.80$\,m, indicating that additional particles yield diminishing returns for the PF but continued benefit for the RBPF.

\begin{figure}[h]
    \centering

    \begin{subfigure}[b]{\linewidth}
    \resizebox{\linewidth}{!}{
    \begin{tikzpicture}
    \begin{axis}[ylabel={RMSE (m)},
            myboxplotstyle,
            xticklabels={},
            ]
    \foreach \particle in {100, 200, 300, 400, 500}{
        \edef\colnameA{RMSE_m\string_0\string_\particle}
        \edef\colnameB{RMSE_m\string_1\string_\particle}
        
        \addplot+ [boxplot] 
            table[y=\colnameA, col sep=comma] {results/adv_diff_benchmark_results.csv};
        \addplot+ [boxplot] 
            table[y=\colnameB, col sep=comma] {results/adv_diff_benchmark_results.csv};
        }
        \end{axis}
        \end{tikzpicture}        }
    \end{subfigure}
    
    \begin{subfigure}[b]{\linewidth}
    \resizebox{\linewidth}{!}{
    \begin{tikzpicture}
    \begin{axis}[ylabel={Final Position Error (m)},
            myboxplotstyle
            ]
    
    \foreach \particle in {100, 200, 300, 400, 500}{
        \edef\colnameA{Final_Error\string_0\string_\particle}
        \edef\colnameB{Final_Error\string_1\string_\particle}
        
        \addplot+ [boxplot] 
            table[y=\colnameA, col sep=comma] {results/adv_diff_benchmark_results.csv};
        \addplot+ [boxplot] 
            table[y=\colnameB, col sep=comma] {results/adv_diff_benchmark_results.csv};
        }
        
        \end{axis}
        \end{tikzpicture}
        }
        \end{subfigure}

    \caption{Comparison of \textcolor{RoyalBlue}{PF} and \textcolor{Orange}{RBPF} localisation accuracy on the
             Advection-Diffusion task across five independent
             runs for varying numbers of particles.
             Interestingly, both metrics have steady RMSE values as the particle count increases. Again, RBPF consistently outperforms PF at equal sample size.}
    \label{fig:pf_rbpf_comparison_adv}
\end{figure}
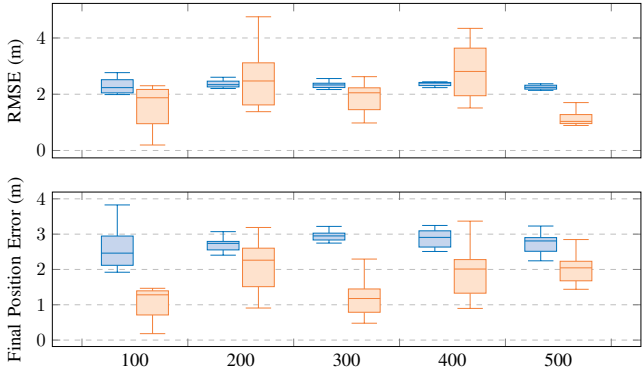

For the advection–diffusion model in figure~\ref{fig:pf_rbpf_comparison_adv}, both filters face a considerably harder localization problem due to the weaker spatial gradients in the scalar concentration field. The PF exhibits remarkably stable but high errors across all particle counts, with RMSE clustered around $2.3$--$2.4$\,m and final errors near $2.2$--$3.0$\,m, suggesting a fundamental accuracy ceiling. The RBPF shows greater variability—its RMSE ranges from $1.23 \pm 0.35$\,m at 500 particles to $3.1 \pm 1.17$\,m at 400 particles—but achieves its best performance at 500 particles, where it reduces RMSE by nearly 47\% relative to the PF ($1.23$\,m vs.\ $2.32$\,m). The high variance across runs reflects the sensitivity of the Kalman update to the weaker and more ambiguous measurement gradients in this environment. These results suggest that the RBPF's analytical correction step offers a clear advantage when sufficient particles are available, but its performance is less predictable than in the richer SWE observation space.

To assess localization sensitivity to multimodal measurement fusion, we conduct a feature ablation study by varying the number of observable state variables. For the SWE model, the full measurement vector comprises three quantities: two velocity components and free-surface elevation, indexed as $0, 1, 2$. For the Advection-Diffusion model, the situation is similar, yet we may choose from a wider range of water quality features: temperature, dissolved oxygen, salinity, turbidity, and chlorophyll. We chose at random a number of measurement channels $ M=1,2,3$, and the Kalman filter dimension, measurement noise, and bias states are reduced accordingly. Five independent runs per configuration capture variability from both random feature selection and stochastic particle dynamics.

\begin{table}[ht]
    \centering
    \resizebox{\linewidth}{!}{
    \begin{tabular}{@{} l cccc @{}}
        \toprule
        & \multicolumn{2}{c}{SWE~\eqref{eq:shallow-water-equation}} & \multicolumn{2}{c}{Advection-Diffsion~\eqref{eq: advection-diffusion-coupled}}\\
        \cmidrule(lr){2-3} \cmidrule(l){4-5}
        $M$ & RMSE (m) & Final Error (m) & RMSE (m) & Final Error (m) \\
        \midrule
        1 & $1.59 \pm 0.55$ & $1.35 \pm 0.49$ & $2.63 \pm 1.6.38$ & $2.11 \pm 0.132$ \\
        2 & $0.94 \pm 0.10$ & $0.90 \pm 0.26$ & $2.27 \pm 1.2.06$ & $1.61 \pm 0.3.9$  \\
        3 & $0.82 \pm 0.06$ & $0.69 \pm 0.08$ & $1.51 \pm 0.79$  & $1.62 \pm 0.64$  \\
        \bottomrule
    \end{tabular}}
    \caption{Performance Comparison of SWE and Advection-Diffusion Models across varying Feature Counts.}
    \label{tab:features}
\end{table}

As shown in Table~\ref{tab:features}, the SWE model degrades gracefully: RMSE increases from $0.82 \pm 0.06$\,m with all three features to $1.59 \pm 0.55$\,m with a single feature—a roughly twofold rise. The advection–diffusion model exhibits markedly steeper degradation, with RMSE growing from $1.51 \pm 0.79$\,m to $2.63 \pm 1.63$\,m and substantially higher variance throughout. These results confirm that fusing multiple field channels as multimodal measurements significantly improves localization accuracy.

\begin{figure}[thbp]
    \centering
    \begin{subfigure}[b]{.59\linewidth}
        \includegraphics[height = 3.0cm]{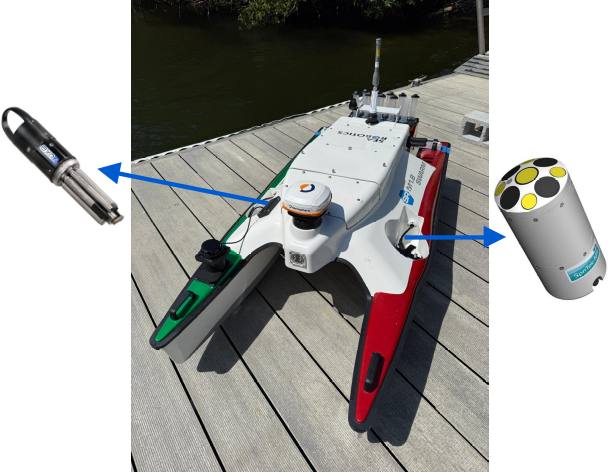}
        \caption{ASV SeaRobotics Suyveyor, left YSI EXO2, right Sontek M9 ADCP.}
        \label{fig:asv}
    \end{subfigure}
    \begin{subfigure}[b]{.39\linewidth}
        \includegraphics[height = 3.0cm]{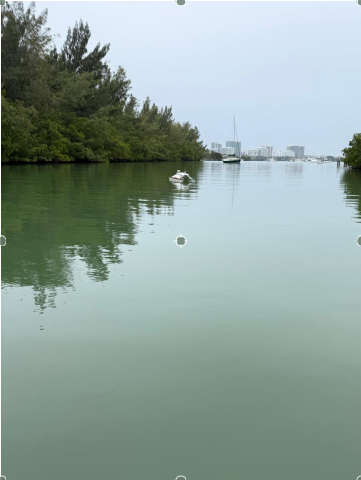}
        \caption{ASV during the field experiment.}
        \label{fig:asv_in_mission}
    \end{subfigure}
    \caption{Hardware utilized during the field experiments; the ASV is deployed in a coastal area where it can use its sensors to measure water quality features, yet it can also measure velocity current speeds, and height from the seafloor. }
    \label{fig:field_experiment}
\end{figure}

\subsection{Physical Experiments}\label{subsec:phisical-experiments}

As shown in Figure~\ref{fig:field_experiment}, we deploy an ASV in a coastal area where it was configured to sense three water quality parameters satisfying the advection diffusion equation~\eqref{eq: advection-diffusion-coupled}: temperature ($^\circ$C), dissolved oxygen ($\text{mg}/\text{L}$), and salinity in Parts Per Thousand (PPT). In the same fashion, the diffusivity constants were set in the same range as in the previous simulations. The velocity field was extracted from the ASV's ADCP.

\begin{figure}[h!]
    \vspace{3pt}
    \centering
    \begin{subfigure}[b]{0.9\linewidth}
        \centering
        \includegraphics[width=\linewidth]{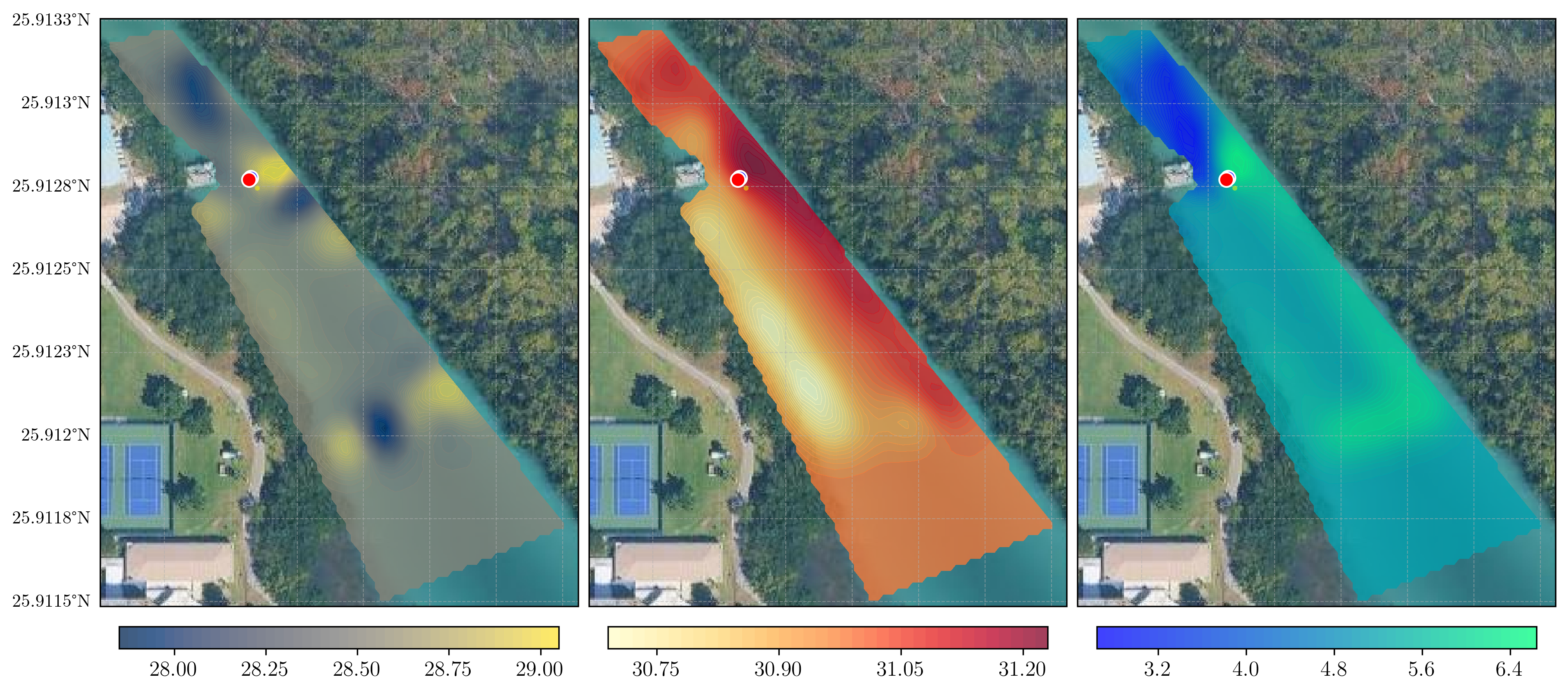}
        \caption{$t = 0$ s}
        \label{fig:adv-diff1}
    \end{subfigure}
    \begin{subfigure}[b]{0.9\linewidth}
        \centering
        \includegraphics[width=\linewidth]{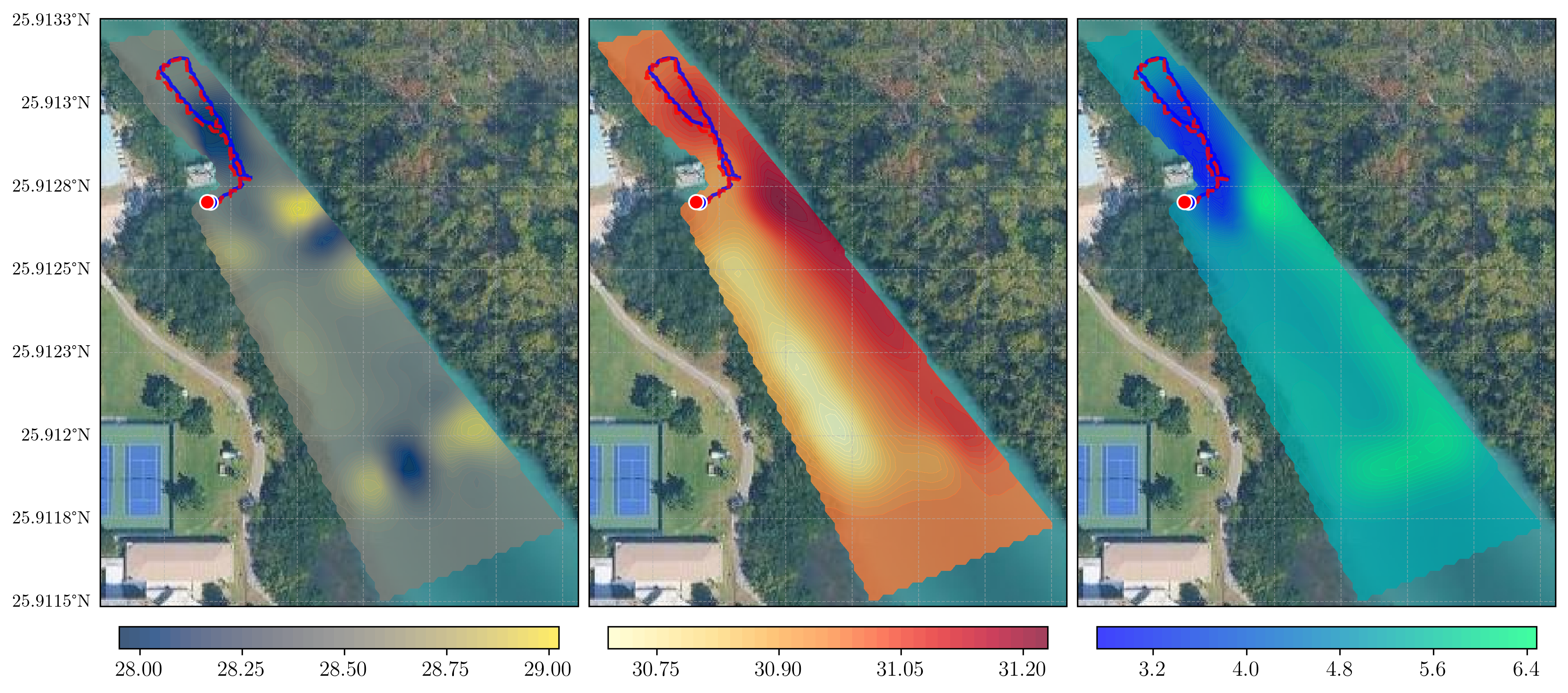}
        \caption{$t = 160$ s}
        \label{fig:adv-diff2}
    \end{subfigure}
    \begin{subfigure}[b]{0.9\linewidth}
        \centering
        \includegraphics[width=\linewidth]{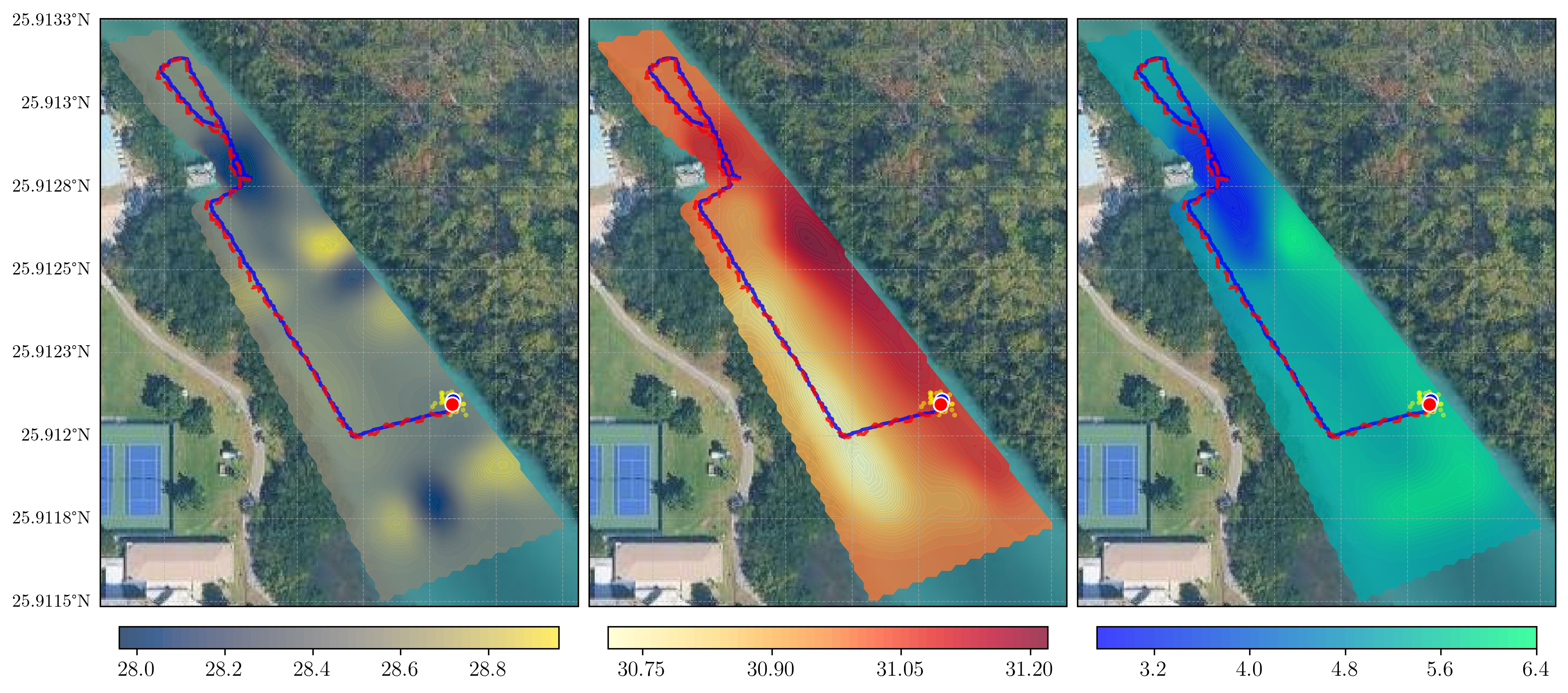}
        \caption{$t = 320$ s}
        \label{fig:adv-diff3}
    \end{subfigure}
    \begin{subfigure}[b]{0.9\linewidth}
        \centering
        \includegraphics[width=\linewidth]{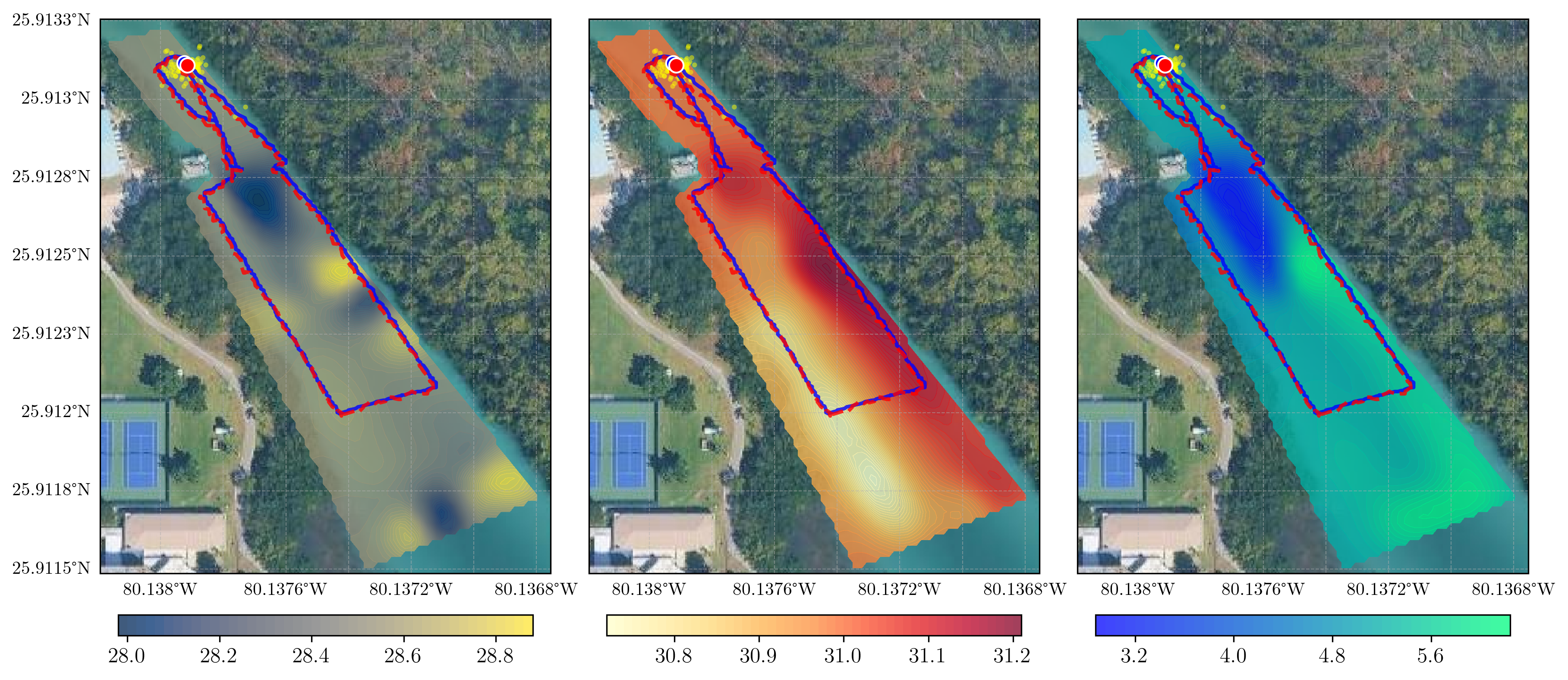}
        \caption{$t = 500$ s}
        \label{fig:adv-diff4}
    \end{subfigure}
    \caption{The path traversed by the ASV, it loops around the region of interest. Again, it uses RBPF localization, measuring $M=3$ water quality features: Salinity (PPT), Temperature ($^\circ$C), and Dissolved oxygen (mgL$^{-1}$).  This shows different timestamps showing the estimation of the robot state over time. The robot estimates the \textcolor{blue}{ground truth path} by estimating its \textcolor{red}{current trajectory} using the \textcolor{Goldenrod}{RBPF} particles.}
    \label{fig:adv-diff-field-experiment}
\end{figure}

The vehicle completed a loop around the designated area, measuring these time-variant variables by measuring the three fields as it moved. The ground truth was obtained by recording the robot's GPS location during its journey. Figure~\ref{fig:adv-diff-field-experiment} shows the predictions made by the RBPF and how the particles are, in general, grouped around the vehicle's true location, which shows that a proper water quality field combination, although dynamic and non-stationary, can provide proper information to extract the vehicle's position with relative precision. 

\section{Conclusion}
This paper presented a localization framework that exploits spatiotemporal environmental fields governed by PDEs as localization signatures for GPS-denied operation. Two PDE classes were formulated, the shallow water equation and the advection-diffusion equation, and fused as multimodal measurements within a Rao-Blackwellized particle filter that maintains closed-form estimates of per-channel sensor bias through conditionally linear Kalman filters.
Simulation results confirm that the RBPF consistently outperforms a standard particle filter in both PDE scenarios, achieving lower RMSE and final position error across all particle counts while requiring significantly fewer particles to reach comparable accuracy. The feature ablation study further demonstrated that fusing multiple field channels as multimodal measurements is critical, with localization accuracy degrading substantially as channels are removed. Field experiments with an ASV measuring salinity, temperature, and dissolved oxygen validated that PDE-governed environmental fields provide sufficient spatial variability for practical localization.
Future work will extend this framework to the SLAM problem, where the vehicle jointly estimates its own state and the underlying environmental field map in real time. Moreover, it will focus on estimating initial conditions, which is a challenge for PDEs, whereas boundary conditions can be derived or assumed from the environment.

\bibliographystyle{ieeetr}
 \begingroup
    \footnotesize
\bibliography{main, references, ref2}

@article{solin2018modeling,
    title={Modeling and interpolation of the ambient magnetic field by {G}aussian processes},
    author={Solin, Arno and Kok, Manon and Wahlstr{\"o}m, Niklas and Sch{\"o}n, Thomas B. and S{\"a}rkk{\"a}, Simo},
    journal={IEEE Transactions on Robotics},
    volume={34},
    number={4},
    pages={1112--1127},
    year={2018},
    publisher={IEEE}
}

@inproceedings{montemerlo2002fastslam,
    title={{FastSLAM}: A factored solution to the simultaneous localization and mapping problem},
    author={Montemerlo, Michael and Thrun, Sebastian and Koller, Daphne and Wegbreit, Ben},
    booktitle={The AAAI Conference on Artificial Intelligence},
      volume={593598},
  number={2},
  pages={593--598},
  year={2002}
}

@inproceedings{solin2016terrain,
    title={Terrain navigation in the magnetic landscape: Particle filtering for indoor positioning},
    author={Solin, Arno and S{\"a}rkk{\"a}, Simo and Kannala, Juho and Rahtu, Esa},
    booktitle={The European Navigation Conference (ENC)},
    pages={1--9},
    year={2016},
    organization={IEEE}
}

@inproceedings{ferris2007wifi,
    title={{WiFi-SLAM} using {G}aussian process latent variable models},
    author={Ferris, Brian and Fox, Dieter and Lawrence, Neil D.},
    booktitle={International Joint Conference on Artificial Intelligence},
    pages={2480--2485},
    year={2007},
    address={Hyderabad, India}
}

@article{dunbabin2012robots,
    title={Robots for environmental monitoring: Significant advancements and applications},
    author={Dunbabin, Matthew and Marques, Lino},
    journal={IEEE Robotics \& Automation Magazine},
    volume={19},
    number={1},
    pages={24--39},
    year={2012},
    publisher={IEEE}
}

@article{doucet2000sequential,
  title={On sequential {M}onte {C}arlo sampling methods for {B}ayesian filtering},
  author={Doucet, Arnaud and Godsill, Simon and Andrieu, Christophe},
  journal={Statistics and Computing},
  volume={10},
  number={3},
  pages={197--208},
  year={2000}
}

@incollection{murphy2001rao,
  title={Rao-{B}lackwellised particle filtering for dynamic {B}ayesian networks},
  author={Murphy, Kevin and Russell, Stuart},
  booktitle={Sequential Monte Carlo Methods in Practice},
  pages={499--515},
  year={2001},
  publisher={Springer}
}

@inproceedings{schon2005marginalized,
  title={Marginalized particle filters for mixed linear/nonlinear state-space models},
  author={Sch{\"o}n, Thomas and Gustafsson, Fredrik and Nordlund, Per-Johan},
  booktitle={IEEE Transactions on Signal Processing},
  volume={53},
  number={7},
  pages={2279--2289},
  year={2005},
  publisher={IEEE}
}

@INPROCEEDINGS{Fuentes2022,
  author={Fuentes, Jose and Bobadilla, Leonardo and Smith, Ryan N.},
  booktitle={IEEE International Conference on Robotic Computing}, 
  title={Localization in Seemingly Sensory-Denied Environments through Spatio-Temporal Varying Fields}, 
  year={2022},
  volume={},
  number={},
  pages={142-147},
  doi={10.1109/IRC55401.2022.00032}}

@inproceedings{song2014towards,
  title={Towards background flow based AUV localization},
  author={Song, Zhuoyuan and Mohseni, Kamran},
  booktitle={IEEE Conference on Decision and Control},
  pages={6945--6950},
  year={2014},
  organization={IEEE}
}

@article{song2019flam,
title={Concurrent flow-based localization and mapping in time-invariant flow fields},
  author={Song, Zhuoyuan and Mohseni, Kamran},
  booktitle={International Conference on Intelligent Robots and Systems},
  pages={7205--7210},
  year={2019},
  organization={IEEE}
}

@inproceedings{kumar2025multiflow,
  title={Flow-based localization and mapping for multi-robot systems},
  author={Kumar, Arjun and Silva, Thales C and Edwards, Victoria and Hsieh, M Ani},
  journal={IEEE Robotics and Automation Letters},
  year={2025},
  publisher={IEEE}
}

@article{li2024enkode,
  title={Enkode: Active learning of unknown flows with koopman operators},
  author={Li, Alice K and Silva, Thales C and Hsieh, M Ani},
  journal={IEEE Robotics and Automation Letters},
  volume={9},
  number={12},
  pages={11282--11289},
  year={2024},
  publisher={IEEE}
}

@inproceedings{fuentes2024adaptive,
   title={Learning-Based Adaptive Navigation for Scalar Field Mapping and Feature Tracking},
  author={Fuentes, Jose and Padr{\~a}o, Paulo and Newaz, Abdullah Al Redwan and Bobadilla, Leonardo},
  booktitle={IEEE International Conference on Robotics and Automation},
  pages={1--7},
  year={2025},
  organization={IEEE}
}

@inproceedings{chen2025mdcpp,
  title={{MDCPP}: Multi-robot Dynamic Coverage Path Planning for Workload Adaptation},
  author={Chen, Jun and Chen, Mingjia and Park, Shinkyu},
  booktitle={arXiv preprint arXiv:2509.23705},
  year={2025}
}

@inproceedings{doucet2000rbpf,
  title={{Rao-Blackwellised} Particle Filtering for Dynamic {Bayesian} Networks},
   author={Murphy, Kevin and Russell, Stuart},
  booktitle={Sequential Monte Carlo methods in practice},
  pages={499--515},
  year={2001},
  publisher={Springer}
}

@article{kok2024rbpf_slam,
    title={Rao-Blackwellized particle smoothing for simultaneous localization and mapping},
  author={Kok, Manon and Solin, Arno and Sch{\"o}n, Thomas B},
  journal={Data-Centric Engineering},
  volume={5},
  pages={e15},
  year={2024},
  publisher={Cambridge University Press}
}

@inproceedings{shen2016connected,
  title={Improving localization accuracy in connected vehicle networks using Rao--Blackwellized particle filters: Theory, simulations, and experiments},
  author={Shen, Macheng and Sun, Jing and Peng, Huei and Zhao, Ding},
  journal={IEEE Transactions on Intelligent Transportation Systems},
  volume={20},
  number={6},
  pages={2255--2266},
  year={2018},
  publisher={IEEE}
}

@inproceedings{taguchi2010rbpf,
   title={Rao-blackwellized particle filtering for probing-based 6-dof localization in robotic assembly},
  author={Taguchi, Yuichi and Marks, Tim K and Okuda, Haruhisa},
  booktitle={IEEE International Conference on Robotics and Automation},
  pages={2610--2617},
  year={2010},
  organization={IEEE}
}

@inproceedings{jouffroy2004underwater,
  title={Underwater vehicle trajectory estimation using contracting PDE-based observers},
  author={Jouffroy, J{\'e}r{\^o}me and Opderbecke, Jan},
  booktitle={American Control Conference},
  volume={5},
  pages={4108--4113},
  year={2004}
}

@inproceedings{wiedemann2017gas,
    title={Probabilistic modeling of gas diffusion with partial differential equations for multi-robot exploration and gas source localization},
  author={Wiedemann, Thomas and Manss, Christoph and Shutin, Dmitriy and Lilienthal, Achim J and Karolj, Valentina and Viseras, Alberto},
  booktitle={European Conference on Mobile Robots},
  pages={1--7},
  year={2017},
  organization={IEEE}
}

@article{newaz2016uav,
  title={UAV-based multiple source localization and contour mapping of radiation fields},
  author={Newaz, Abdullah Al Redwan and Jeong, Sungmoon and Lee, Hosun and Ryu, Hyejeong and Chong, Nak Young},
  journal={Robotics and Autonomous Systems},
  volume={85},
  pages={12--25},
  year={2016},
  publisher={Elsevier}
}

@article{teixeira2016auv,
  title={AUV terrain-aided navigation using a Doppler velocity logger},
  author={Teixeira, Francisco Curado and Quintas, Jo{\~a}o and Pascoal, Ant{\'o}nio},
  journal={Annual Reviews in Control},
  volume={42},
  pages={166--176},
  year={2016},
  publisher={Elsevier}
}

@article{mcconnell2022perception,
  title={Perception for underwater robots},
  author={McConnell, John and Collado-Gonzalez, Ivana and Englot, Brendan},
  journal={Current Robotics Reports},
  volume={3},
  number={4},
  pages={177--186},
  year={2022},
  publisher={Springer}
}

@incollection{islam2024computer,
  title={Computer vision applications in underwater robotics and oceanography},
  author={Islam, Md Jahidul and Li, Alberto Quattrini and Girdhar, Yogesh A and Rekleitis, Ioannis},
  booktitle={Computer Vision},
  pages={173--204},
  year={2024},
  publisher={Chapman and Hall/CRC}
}
\end{document}